\documentclass[10pt]{paper}

\usepackage{microtype}
\usepackage{algorithm}
\usepackage{algorithmic}
\usepackage{pgfplots}
\pgfplotsset{compat=1.18}

\newcommand{\R}{\mathbb{R}}
\newcommand{\loss}{\mathcal{L}}
\newcommand{\dataset}{\mathcal{D}}
\newcommand{\safety}{\mathcal{S}}
\newcommand{\task}{\mathcal{T}}

\DeclareMathOperator{\tr}{tr}

\begin{document}
\begin{center}

\title{SafeAnchor: Preventing Cumulative Safety Erosion\\in Continual Domain Adaptation of Large Language Models}

\maketitle

\thispagestyle{empty}
\pagenumbering{gobble}

% Camera-ready: de-anonymized authors
\begin{tabular}{c}
	Dongxin Guo\upstairs{\affilone}\quad
	Jikun Wu\upstairs{\affiltwo}\quad
	Siu Ming Yiu\upstairs{\affilone}
	\\[0.6ex]
	{\small\upstairs{\affilone}\,The University of Hong Kong, Hong Kong, China}
	\\[0.1ex]
	{\small\upstairs{\affiltwo}\,Brain Investing Limited, Hong Kong, China}
	\\
\end{tabular}

\vspace*{0.1in}
\end{center}

\begin{abstract}
	Safety alignment in large language models is remarkably shallow: it is concentrated in the first few output tokens and reversible by fine-tuning on as few as 100 adversarial examples. This fragility becomes critical in real-world deployment, where models undergo sequential adaptation across domains such as medicine, law, and code, causing safety guardrails to erode cumulatively. Yet all existing safety-preserving methods target only single-task fine-tuning, leaving the multi-domain sequential setting entirely unaddressed.
	
	We introduce \textbf{SafeAnchor}, a framework that anchors safety in place throughout continual adaptation. SafeAnchor first identifies low-rank safety subspaces in LoRA parameter space via Fisher Information eigendecomposition, then constrains domain-specific gradient updates to the orthogonal complement of these subspaces, and finally monitors for residual safety drift with threshold-triggered corrective replay. Evaluated on Llama-2-7B-Chat and Mistral-7B-Instruct across a three-domain pipeline and eight benchmarks, SafeAnchor retains \textbf{93.2\%} of original safety alignment, outperforming all baselines by 18--42 points, while matching unconstrained fine-tuning to within 1.5 points on domain tasks.
\end{abstract}


%%% SECTION 1: INTRODUCTION %%%
\section{Introduction}
\label{sec:intro}

Large language models (LLMs) are increasingly adapted to specialized domains such as medicine, law, and code generation through continual fine-tuning on domain-specific data \cite{shi2025clsurvey, luo2025forgetting}. In practice, a single model may undergo a sequence of adaptations as organizational needs evolve, forming a \emph{continual domain adaptation} pipeline. While this sequential fine-tuning extends model capabilities, it poses a critical yet underexplored risk: the cumulative erosion of safety alignment.

Recent work has established that safety alignment in LLMs is surprisingly brittle. Qi et al.~\cite{qi2025safety} demonstrated that current alignment primarily modifies the generative distribution over the first few output tokens, teaching models to begin with refusal phrases rather than embedding deep behavioral changes. Ji et al.~\cite{ji2025resist} introduced \emph{alignment elasticity}, showing that LLMs inherently resist alignment and revert to pre-training behavior upon further fine-tuning, with this effect intensifying at larger scales. Even benign fine-tuning on standard datasets can compromise safety guardrails \cite{qi2024finetuning}, and as few as 100 adversarial examples can undo 100,000 safety training instances \cite{yang2023shadow}. The ``safety tax'' documented by Huang et al.~\cite{huang2025safetytax} further reveals a fundamental trade-off between safety and reasoning capability.

These findings reveal a fundamental vulnerability: each adaptation step risks degrading safety, and these degradations compound. Yet existing safety-preserving methods (Vaccine \cite{huang2024vaccine}, RepNoise \cite{rosati2024repnoise}, Safe LoRA \cite{hsu2024safelora}, SaLoRA \cite{li2025salora}, Lisa \cite{huang2024lisa}, SafeGrad \cite{yi2025safegrad}) all address only single-task adaptation. CL methods such as O-LoRA \cite{wang2023olora}, InfLoRA \cite{liang2024inflora}, and GainLoRA \cite{liang2025gainlora} prevent task forgetting but ignore safety. Alssum et al.~\cite{alssum2024unforgotten} and OGPSA \cite{ogpsa2026} evaluate only single-task or the inverse problem. No existing work addresses safety during \emph{sequential multi-domain} continual adaptation.

We propose \textbf{SafeAnchor}, a framework that anchors safety alignment in place while enabling effective continual domain adaptation of LLMs. SafeAnchor integrates three components whose combination addresses the sequential safety-preservation setting (Figure~\ref{fig:framework}):

\textbf{(1) Safety Subspace Identification (SSI)} uses gradient-based Fisher Information analysis \cite{arditi2024refusal, wollschlager2025hidden, zou2023representation} to identify LoRA parameter directions encoding safety, updated incrementally after each domain via explicit subspace merging.

\textbf{(2) Orthogonal Safety-Constrained Adaptation (OSCA)} projects LoRA gradient updates onto the orthogonal complement of the safety subspace, extending gradient projection from single-task safety \cite{yi2025safegrad} and continual learning \cite{liang2024inflora, farajtabar2020orthogonal} to the sequential safety-preservation setting.

\textbf{(3) Cumulative Safety Monitoring (CSM)} evaluates safety on a held-out probe set after each domain using LlamaGuard \cite{inan2023llamaguard}, triggering corrective replay when degradation exceeds a threshold.

We evaluate on Llama-2-7B-Chat \cite{touvron2023llama} and Mistral-7B-Instruct \cite{jiang2023mistral} across eight benchmarks: domain (MedQA \cite{jin2020medqa}, LegalBench \cite{guha2024legalbench}, HumanEval \cite{chen2021humaneval}), safety (HarmBench \cite{mazeika2024harmbench}, TruthfulQA \cite{lin2022truthfulqa}, BBQ \cite{parrish2022bbq}, WildGuard \cite{han2024wildguard}), and general (MMLU \cite{hendrycks2021mmlu}). All results report mean $\pm$ std over five seeds. SafeAnchor retains $93.2\pm1.0$\% of original safety within 1.5 points of unconstrained fine-tuning, outperforming all baselines by 18--42 points.

Our contributions are:
\begin{itemize}[leftmargin=*,nosep]
\item We identify and characterize the problem of \emph{safety-preserving continual domain adaptation}, demonstrating that sequential fine-tuning causes compounding safety erosion unaddressed by existing methods.
\item We propose SafeAnchor, integrating Fisher-based subspace identification, orthogonal gradient projection, and threshold-triggered monitoring into a unified framework for sequential safety preservation.
\item We provide comprehensive experiments on two models and eight benchmarks with variance reporting, ablations, and sensitivity analysis, achieving state-of-the-art safety preservation.
\end{itemize}

%%% SECTION 2: RELATED WORK %%%
%%% SECTION 2: RELATED WORK (OPTIMIZED) %%%
\section{Related Work}
\label{sec:related}

\subsection{Fragility of Safety Alignment}
\label{sec:related_fragility}

Standard alignment techniques produce safety behaviors that are surprisingly shallow. Qi et al.~\cite{qi2025safety} showed alignment concentrates in the first few output tokens; Ji et al.~\cite{ji2025resist} demonstrated that LLMs exhibit \emph{alignment elasticity}, reverting to pre-training behavior upon further fine-tuning, with this effect worsening at scale. Even benign fine-tuning compromises safety \cite{qi2024finetuning}, and as few as 100 adversarial examples can undo extensive safety training \cite{yang2023shadow}. The safety tax \cite{huang2025safetytax} further documents a fundamental safety--capability trade-off. Together, these findings motivate our work: if single-step fine-tuning already degrades safety, sequential multi-domain adaptation compounds the problem.

\subsection{Safety-Preserving Fine-Tuning}
\label{sec:related_safety}

Existing defenses operate at three stages \cite{hft_survey2024}. \emph{Alignment-stage} methods modify the base model before fine-tuning: Vaccine \cite{huang2024vaccine} produces perturbation-invariant embeddings, and RepNoise \cite{rosati2024repnoise} removes harmful representations. \emph{Fine-tuning-stage} methods constrain the adaptation process: Lisa \cite{huang2024lisa} alternates alignment and task optimization, and SafeGrad \cite{yi2025safegrad} projects away safety-conflicting gradients. \emph{Post-fine-tuning} methods adjust adapted weights: Safe LoRA \cite{hsu2024safelora} projects LoRA onto safety-aligned subspaces, and SaLoRA \cite{li2025salora} orthogonalizes against safety features. Despite this rich landscape, \textbf{every method addresses only single-task adaptation}, leaving cumulative erosion across sequential domains unaddressed.

\subsection{Continual Learning, Orthogonal Gradients, and Safety Subspaces}
\label{sec:related_cl}

Continual learning for LLMs has converged on LoRA-based approaches \cite{hu2022lora, shi2025clsurvey}. Orthogonal subspace methods are most relevant to our work: O-LoRA \cite{wang2023olora} learns each task in a subspace orthogonal to the column span of prior-task LoRA matrices, preventing inter-task interference but providing no mechanism for protecting a behavioral property; InfLoRA \cite{liang2024inflora} pre-allocates per-task directions via SVD of input activations; OGD \cite{farajtabar2020orthogonal} provides the theoretical framing for gradient projection which we extend from task-preservation to safety-preservation; and GainLoRA \cite{liang2025gainlora} uses input-dependent routing, an orthogonal design axis. EWC \cite{kirkpatrick2017ewc} penalizes changes to parameters important for prior tasks via diagonal Fisher, which inspired our Fisher-based subspace identification, though we apply it to safety-behavior preservation rather than task-performance.

Separately, mechanistic interpretability has revealed that safety occupies identifiable low-rank subspaces. Arditi et al.~\cite{arditi2024refusal} found refusal is mediated by a single activation direction; Wollschl\"ager et al.~\cite{wollschlager2025hidden} extended this to multi-dimensional spaces; representation engineering \cite{zou2023representation} provides broader frameworks; DARE \cite{yu2024dare} reveals redundancy in LoRA deltas; and SAILS \cite{sails2024} constructs interpretable safety subspaces via sparse autoencoders.

Two works partially bridge CL and safety: OGPSA \cite{ogpsa2026} addresses the \emph{inverse} problem of preserving task knowledge during safety alignment, and Alssum et al.~\cite{alssum2024unforgotten} apply CL methods to safety but evaluate only single-task settings. SafeAnchor is the first to combine safety subspace identification with orthogonally-constrained continual LoRA optimization across \emph{sequential multi-domain} adaptation.

%%% SECTION 3: METHOD %%%
\section{SafeAnchor}
\label{sec:method}

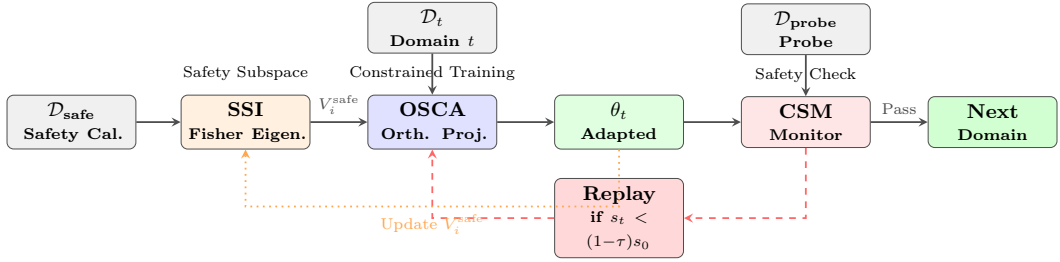
\begin{figure}[t]
\centering
\resizebox{\columnwidth}{!}{%
\begin{tikzpicture}[
    node distance=0.6cm and 0.8cm,
    box/.style={rectangle, draw=black!70, fill=#1, rounded corners=3pt, minimum height=0.7cm, minimum width=1.8cm, font=\small\bfseries, align=center, text width=1.8cm},
    box/.default=blue!8,
    arrow/.style={->, >=stealth, thick, color=black!70},
    label/.style={font=\scriptsize, align=center},
]

% SSI Block
\node[box=orange!12] (ssi) {SSI\\[-1pt]{\scriptsize Fisher Eigen.}};
\node[label, above=0.1cm of ssi] {Safety Subspace};

% Safety set
\node[box=gray!12, left=0.7cm of ssi, minimum width=1.2cm] (safeset) {$\dataset_{\text{safe}}$\\[-1pt]{\scriptsize Safety Cal.}};

% OSCA Block
\node[box=blue!12, right=0.9cm of ssi] (osca) {OSCA\\[-1pt]{\scriptsize Orth. Proj.}};
\node[label, above=0.1cm of osca] {Constrained Training};

% Domain data
\node[box=gray!12, above=0.65cm of osca, minimum width=1.2cm] (domdata) {$\dataset_t$\\[-1pt]{\scriptsize Domain $t$}};

% Adapted model
\node[box=green!12, right=0.9cm of osca] (adapted) {$\theta_t$\\[-1pt]{\scriptsize Adapted}};

% CSM Block
\node[box=red!10, right=0.9cm of adapted] (csm) {CSM\\[-1pt]{\scriptsize Monitor}};
\node[label, above=0.1cm of csm] {Safety Check};

% Probe set
\node[box=gray!12, above=0.65cm of csm, minimum width=1.2cm] (probe) {$\dataset_{\text{probe}}$\\[-1pt]{\scriptsize Probe}};

% Output
\node[box=green!18, right=0.9cm of csm] (output) {Next\\[-1pt]{\scriptsize Domain}};

% Replay arc
\node[box=red!15, below=0.45cm of adapted, minimum width=1.6cm] (replay) {Replay\\[-1pt]{\scriptsize if $s_t < (1{-}\tau)s_0$}};

% Arrows
\draw[arrow] (safeset) -- (ssi);
\draw[arrow] (ssi) -- node[above, label] {$V_i^{\text{safe}}$} (osca);
\draw[arrow] (domdata) -- (osca);
\draw[arrow] (osca) -- (adapted);
\draw[arrow] (adapted) -- (csm);
\draw[arrow] (probe) -- (csm);
\draw[arrow] (csm) -- node[above, label] {Pass} (output);
\draw[arrow, dashed, red!60] (csm) |- (replay);
\draw[arrow, dashed, red!60] (replay) -| (osca);

% Update arc (back from adapted to SSI)
\draw[arrow, dotted, orange!70, thick] (adapted.south) -- ++(0,-0.9) -| node[below, label, pos=0.25] {Update $V_i^{\text{safe}}$} (ssi.south);

\end{tikzpicture}%
}
\caption{SafeAnchor pipeline. SSI identifies safety-critical LoRA directions via Fisher eigendecomposition. OSCA projects domain gradients orthogonally during training. CSM triggers corrective replay if safety degrades. The subspace is incrementally updated after each domain.}
\label{fig:framework}
\end{figure}

We present SafeAnchor, a framework for preserving safety alignment during continual domain adaptation of LLMs. Figure~\ref{fig:framework} illustrates the overall pipeline and Algorithm~\ref{alg:safeanchor} summarizes the procedure.

% === FRAMEWORK OVERVIEW FIGURE (TikZ) ===

\subsection{Problem Formulation}
\label{sec:problem}

Consider a safety-aligned LLM with parameters $\theta_0$ that must be sequentially adapted to $T$ domains $\{\task_1, \task_2, \ldots, \task_T\}$, each with training data $\dataset_t$. Using LoRA \cite{hu2022lora}, the model's weight matrices $W \in \R^{d \times k}$ are augmented as $W + BA$, where $B \in \R^{d \times r}$ and $A \in \R^{r \times k}$ with rank $r \ll \min(d, k)$. At each step $t$, we train LoRA parameters $\Delta_t = \{B_t, A_t\}$.

Let $\safety(\theta)$ denote the safety score and $P_t(\theta)$ denote domain task performance. The standard CL objective maximizes $\sum_t P_t(\theta_t)$ subject to preventing forgetting. \textbf{Our objective adds a safety constraint}: find $\{\Delta_t\}_{t=1}^T$ that maximizes domain performance while maintaining $\safety(\theta_t) \geq (1-\epsilon)\safety(\theta_0)$ for all $t$, where $\epsilon$ is a small tolerance. As we demonstrate, this is highly non-trivial without targeted intervention.

\subsection{Safety Subspace Identification (SSI)}
\label{sec:ssi}

Inspired by findings that safety behaviors occupy identifiable low-rank subspaces in both activation space \cite{arditi2024refusal, wollschlager2025hidden, zou2023representation} and parameter space \cite{yu2024dare, sails2024}, we identify the parameter subspace encoding safety-critical behavior using a gradient-based Fisher Information approach.

\textbf{Safety Calibration Set.} We construct a small calibration set $\dataset_{\text{safe}}$ of $N_s$ prompt-response pairs (we use $N_s = 500$) drawn from a safety-aligned dataset \cite{dai2024safe, ji2024beavertails}, containing both harmful prompts with correct refusal responses and benign prompts with helpful responses.

\textbf{Fisher Information for LoRA Parameters.} For each LoRA layer $i$, we denote its flattened parameter vector as $\delta_i = \text{vec}([B_i; A_i])$, concatenating the vectorized $B_i$ and $A_i$ matrices from the set-level LoRA parameters $\Delta_t$. We compute the empirical Fisher Information Matrix:
\begin{equation}
\label{eq:fisher}
F_i = \frac{1}{N_s} \sum_{(x,y) \in \dataset_{\text{safe}}} \nabla_{\delta_i} \log p_\theta(y|x) \, \nabla_{\delta_i} \log p_\theta(y|x)^\top
\end{equation}
We acknowledge that the empirical Fisher is a biased approximation of the true Fisher; we verify robustness to $N_s$ in Section~\ref{sec:sensitivity}. The eigendecomposition $F_i = U_i \Lambda_i U_i^\top$ reveals principal safety-relevant directions. We select eigenvectors whose eigenvalues account for cumulative proportion $\rho$ of total variance ($\rho=90\%$) to form the \emph{safety subspace basis}:
\begin{equation}
\label{eq:safety_subspace}
V_i^{\text{safe}} = [u_i^{(1)}, u_i^{(2)}, \ldots, u_i^{(k_s)}] \in \R^{|\delta_i| \times k_s}
\end{equation}
The resulting projection matrix for layer $i$ is $\Pi_i^{\text{safe}} = V_i^{\text{safe}} (V_i^{\text{safe}})^\top$. As we verify in Section~\ref{sec:ablation}, the eigenvalue spectrum is sharply decaying, confirming that safety information occupies a genuinely low-rank subspace in LoRA parameter space, consistent with activation-space findings \cite{arditi2024refusal}.

\textbf{Incremental Subspace Update.} After each domain adaptation $t$, the model parameters shift, potentially altering which directions are safety-critical. We update the safety subspace as follows: (1)~recompute $F_i$ on $\dataset_{\text{safe}}$ using the adapted model $\theta_t$ to obtain new eigenvectors $\hat{V}_i^{\text{safe}}$; (2)~concatenate the old and new bases: $V_i^{\text{merged}} = [V_i^{\text{safe}} \mid \hat{V}_i^{\text{safe}}]$; (3)~perform SVD on $V_i^{\text{merged}}$ and retain the top singular vectors capturing $\rho$ of the variance, yielding the updated $V_i^{\text{safe}}$ with controlled rank. This ensures the safety subspace tracks the evolving model while preventing unbounded rank growth. Note that SVD truncation may discard old safety directions with diminished singular values; we verify in Section~\ref{sec:ablation} that the resulting subspace remains effective.

\subsection{Orthogonal Safety-Constrained Adaptation (OSCA)}
\label{sec:osca}

For domain $t$, OSCA constrains LoRA updates to lie in the orthogonal complement of the safety subspace. Given a task loss gradient $g_i^t = \nabla_{\delta_i} \loss_t$ at layer $i$, OSCA computes the projected gradient:
\begin{equation}
\label{eq:projection}
\tilde{g}_i^t = g_i^t - \Pi_i^{\text{safe}} \, g_i^t = (I - V_i^{\text{safe}} (V_i^{\text{safe}})^\top) \, g_i^t
\end{equation}
This removes the safety-subspace component from the gradient, ensuring domain-specific learning occurs only in directions orthogonal to safety-critical parameters \cite{farajtabar2020orthogonal}.

\textbf{Adaptive Projection Strength.} Strict orthogonal projection may over-constrain learning on layers where safety and task subspaces overlap significantly. We introduce an adaptive relaxation coefficient $\alpha_i \in [0, 1]$:
\begin{equation}
\label{eq:adaptive}
\hat{g}_i^t = \tilde{g}_i^t + \alpha_i \cdot \Pi_i^{\text{safe}} \, g_i^t
\end{equation}
where $\alpha_i = \max(0, 1 - \lambda \cdot \tr(F_i))$ decays toward zero for layers with high safety importance (large Fisher trace $\tr(F_i) = \sum_j \mu_j^{(i)}$, the sum of eigenvalues), enforcing stricter projection where safety is most concentrated. Here $\lambda$ is a hyperparameter controlling the strictness.

\subsection{Cumulative Safety Monitoring (CSM)}
\label{sec:csm}

While OSCA prevents direct interference with the safety subspace, indirect effects (such as changes to shared representations that affect safety through non-linear pathways) may still accumulate across domain transitions. CSM addresses this through a lightweight monitoring and correction mechanism.

\textbf{Safety Probe.} We maintain a held-out safety probe set $\dataset_{\text{probe}}$ (disjoint from $\dataset_{\text{safe}}$, 200 examples from HarmBench \cite{mazeika2024harmbench}). After each domain adaptation $t$, we evaluate the safety refusal rate $s_t$ on $\dataset_{\text{probe}}$ using LlamaGuard \cite{inan2023llamaguard} as the binary safety classifier (which achieves 92.1\% F1 on our probe set for distinguishing safe refusals from harmful completions).

\textbf{Threshold-Triggered Safety Replay.} If $s_t < (1-\tau)\, s_0$ (where $s_0$ is the baseline refusal rate and $\tau$ is the fractional tolerance), CSM triggers a brief safety replay: we fine-tune for $E_{\text{repair}}=200$ steps on a mixture of the safety calibration set and the current domain data, using the OSCA-projected gradients, to restore the safety score. The value $E_{\text{repair}}=200$ sits at the knee of a diminishing-returns curve (Appendix~\ref{app:csm}): $100$ steps recover only $+1.9$ points vs. $+2.7$ at $200$, while $400$ adds a further $+0.2$ points at twice the cost plus $0.4$ points of domain regression. After replay, we re-evaluate $s_t$ to verify $s_t \geq (1-\tau)\, s_0$; if not, replay is extended by at most one further $E_{\text{repair}}$ block. The replay objective is:
\begin{equation}
\label{eq:replay}
\loss_{\text{replay}} = \loss_t(\dataset_t) + \beta \cdot \loss_{\text{safe}}(\dataset_{\text{safe}})
\end{equation}
where $\beta$ controls the safety-task balance. Empirically, CSM triggers $3$ times across $5$ seeds $\times$ $3$ domains ($0.20$ triggers per domain-adaptation, concentrated on Code; Appendix~\ref{app:csm}), adding negligible overhead.

\subsection{Training Objective}
\label{sec:training}

The complete SafeAnchor training objective for domain $t$ combines the domain task loss with a safety regularization term:
\begin{equation}
\label{eq:total_loss}
\loss_{\text{total}}^t = \loss_t(\dataset_t) + \gamma \cdot \loss_{\text{anchor}}^t
\end{equation}
where the \emph{safety anchor loss} penalizes changes to the model's output distribution on safety-relevant inputs:
\begin{equation}
\label{eq:anchor}
\loss_{\text{anchor}}^t = \frac{1}{|\dataset_{\text{safe}}|} \sum_{x \in \dataset_{\text{safe}}} D_{\text{KL}}\big(p_{\theta_{t-1}}(\cdot|x) \,\|\, p_{\theta_t}(\cdot|x)\big)
\end{equation}
We use forward KL because it is mean-seeking: it penalizes the current model for assigning low probability where the safe model assigned high probability, preserving refusal behaviors. Empirically, reverse KL yields 1.8 points lower safety. The coefficient $\gamma$ controls regularization strength. This anchor loss complements OSCA's hard orthogonal projection with soft distributional regularization. Note that OSCA projects only the task gradient $\nabla \loss_t$, while the anchor gradient $\nabla \loss_{\text{anchor}}^t$ bypasses projection so that its safety-reinforcing signal is preserved.

\begin{algorithm}[t]
\caption{SafeAnchor: Safety-Preserving Continual Domain Adaptation}
\label{alg:safeanchor}
\begin{algorithmic}[1]
\REQUIRE Safety-aligned model $\theta_0$, domains $\{\task_1, \ldots, \task_T\}$, safety sets $\dataset_{\text{safe}}, \dataset_{\text{probe}}$, LlamaGuard classifier $C$
\STATE Compute initial safety subspace $\{V_i^{\text{safe}}\}$ via SSI (Eq.~\ref{eq:fisher}--\ref{eq:safety_subspace})
\STATE Evaluate baseline safety $s_0 = C(\theta_0, \dataset_{\text{probe}})$
\FOR{$t = 1$ to $T$}
    \STATE Initialize LoRA parameters $\Delta_t$ for domain $\task_t$
    \FOR{each training step}
        \STATE Compute task gradient $g_i^t \gets \nabla_{\delta_i} \loss_t$; anchor gradient $a_i^t \gets \gamma\nabla_{\delta_i}\loss_{\text{anchor}}^t$
        \STATE Apply OSCA: $\hat{g}_i^t \gets \tilde{g}_i^t + \alpha_i \cdot \Pi_i^{\text{safe}} g_i^t$ \COMMENT{Eqs.~\ref{eq:projection}--\ref{eq:adaptive}}
        \STATE Update $\delta_i \gets \delta_i - \eta (\hat{g}_i^t + a_i^t)$
    \ENDFOR
    \STATE Evaluate $s_t = C(\theta_t, \dataset_{\text{probe}})$ using LlamaGuard
    \IF{$s_t < (1-\tau)\, s_0$}
        \STATE Run safety replay (Eq.~\ref{eq:replay}) for $E_{\text{repair}}$ steps; re-evaluate $s_t$; extend by at most one further block if still below threshold
    \ENDIF
    \STATE Recompute $\hat{V}_i^{\text{safe}}$ from $\theta_t$; merge with $V_i^{\text{safe}}$ via SVD truncation at $\rho$
\ENDFOR
\RETURN Adapted model $\theta_T$
\end{algorithmic}
\end{algorithm}

%%% SECTION 4: EXPERIMENTS %%%
\section{Experiments}
\label{sec:experiments}

\subsection{Experimental Setup}
\label{sec:setup}

\textbf{Models.} We evaluate on Llama-2-7B-Chat \cite{touvron2023llama} and Mistral-7B-Instruct \cite{jiang2023mistral}, both safety-aligned chat models representative of widely deployed open-weight LLMs.

\textbf{Sequential Domain Adaptation.} We adapt models through three domains in sequence: (1)~\emph{Medical}: fine-tuned on MedQA training split \cite{jin2020medqa} for medical question answering; (2)~\emph{Legal}: fine-tuned on LegalBench tasks \cite{guha2024legalbench} for legal reasoning; (3)~\emph{Code}: fine-tuned on the CodeAlpaca subset for code generation. Each domain uses 5,000 training examples with 3 epochs. We additionally test an alternative ordering (Code $\rightarrow$ Legal $\rightarrow$ Medical) in Section~\ref{sec:robustness}.

\textbf{Evaluation Benchmarks.} \emph{Safety metrics:} HarmBench \cite{mazeika2024harmbench} (refusal rate on 200 harmful prompts), TruthfulQA \cite{lin2022truthfulqa} (truthfulness score), BBQ \cite{parrish2022bbq} (bias score; lower is better, inverted for composite), and WildGuard \cite{han2024wildguard} (jailbreak robustness). We compute a composite \emph{Safety Score} as:
\begin{equation}
\label{eq:safetyscore}
\text{Safety} = \frac{1}{3}\!\left(\frac{\text{HarmBench}}{100} + \frac{\text{TruthfulQA}}{100} + \frac{100-\text{BBQ}_{\text{bias}}}{100}\right) \times 100
\end{equation}
where each component is expressed as a percentage and BBQ bias is inverted so higher is better. WildGuard is reported separately as an independent jailbreak-robustness indicator distinct from the refusal/truthfulness/bias triad. \emph{Domain metrics:} MedQA accuracy, LegalBench accuracy, and HumanEval pass@1 \cite{chen2021humaneval}. \emph{General:} MMLU \cite{hendrycks2021mmlu}. MT-Bench \cite{zheng2023mtbench} scores are reported in Section~\ref{sec:robustness}.

\textbf{Baselines.} (1)~\emph{Standard LoRA} \cite{hu2022lora}: unconstrained sequential fine-tuning. (2)~\emph{EWC+LoRA} \cite{kirkpatrick2017ewc}: Fisher-based regularization. (3)~\emph{O-LoRA} \cite{wang2023olora}: orthogonal subspace LoRA. (4)~\emph{Safe LoRA} \cite{hsu2024safelora}: post-hoc safety projection. (5)~\emph{Vaccine+LoRA} \cite{huang2024vaccine}: pre-immunization. (6)~\emph{SafeGrad+LoRA} \cite{yi2025safegrad}: gradient surgery. (7)~\emph{Safety Interleaving}: mixing 10\% safety data from BeaverTails into each domain's training set (a natural but previously untested baseline).

\textbf{Baseline Adaptation to Sequential Setting.} For fair comparison, all baselines were adapted to the sequential pipeline: SafeGrad's alignment gradient is recomputed per domain step; Safe LoRA's projection is applied after each step; Vaccine's immunization is applied once before all domains (per its design); EWC's Fisher is recomputed per domain; O-LoRA allocates orthogonal subspaces per domain. All use identical LoRA configurations.

\textbf{Implementation.} PEFT \cite{mangrulkar2022peft} with LoRA $r{=}16$, $\alpha{=}32$ on $Q,K,V,O$ projections. Learning rate $2{\times}10^{-4}$ (cosine), batch 8 $\times$ grad-accum 2, AdamW ($\beta_1{=}0.9, \beta_2{=}0.999,$ wd$=0.01$). Safety calibration: 500 BeaverTails examples \cite{ji2024beavertails}. Defaults: $\rho{=}0.90, \tau{=}0.05, \gamma{=}0.1, \lambda{=}0.5, \beta{=}1.0, E_{\text{repair}}{=}200$. All runs on 2$\times$A100 40GB (BF16). A 3-domain, 5-seed pipeline takes ${\approx}7$\,h\,47\,min per seed: SSI ${\approx}12.4$\,min/domain, OSCA ${\approx}17.8\%$ per-step overhead, CSM probe ${\approx}4.8$\,min/transition (plus ${\approx}5$\,min replay if triggered). Mean $\pm$ std over 5 seeds throughout. Complete hyperparameters and per-phase compute in Appendix~\ref{app:hparams}.

\subsection{Main Results}
\label{sec:results}

\begin{table}[t]
\centering
\caption{Results on Llama-2-7B-Chat after sequential adaptation (Medical $\rightarrow$ Legal $\rightarrow$ Code). Safety Score is the composite of HarmBench, TruthfulQA, and BBQ (Eq.~\ref{eq:safetyscore}). All values: mean $\pm$ std over 5 seeds. Best in \textbf{bold}, second \underline{underlined}.}
\label{tab:main}
\begin{tabular}{l c c c}
\toprule
\textbf{Method} & \textbf{Safety$\uparrow$} & \textbf{Domain$\uparrow$} & \textbf{MMLU$\uparrow$} \\
\midrule
Base (no adapt.) & $91.4$ & $38.2$ & $46.1$ \\
\midrule
Standard LoRA & $43.6 \pm 2.1$ & $\mathbf{62.7 \pm 0.6}$ & $44.8 \pm 0.3$ \\
EWC + LoRA & $52.1 \pm 1.8$ & $59.4 \pm 0.7$ & $45.3 \pm 0.4$ \\
O-LoRA & $48.7 \pm 2.3$ & $60.1 \pm 0.8$ & $45.0 \pm 0.3$ \\
Safe LoRA & $61.3 \pm 1.5$ & $57.8 \pm 0.9$ & $44.9 \pm 0.4$ \\
Vaccine + LoRA & $58.9 \pm 1.7$ & $58.6 \pm 0.8$ & $44.5 \pm 0.5$ \\
SafeGrad + LoRA & $67.4 \pm 1.4$ & $59.2 \pm 0.7$ & $45.1 \pm 0.3$ \\
Safety Interleaving & $64.8 \pm 1.6$ & $60.9 \pm 0.7$ & $45.2 \pm 0.4$ \\
\midrule
\textbf{SafeAnchor} & $\mathbf{85.2 \pm 0.9}$ & $\underline{61.4 \pm 0.5}$ & $\mathbf{45.7 \pm 0.3}$ \\
\bottomrule
\end{tabular}
\end{table}

\begin{table}[t]
\centering
\caption{Results on Mistral-7B-Instruct across the same sequential pipeline. SafeAnchor demonstrates consistent safety preservation across architectures. All values: mean $\pm$ std over 5 seeds.}
\label{tab:mistral}
\begin{tabular}{l c c c}
\toprule
\textbf{Method} & \textbf{Safety$\uparrow$} & \textbf{Domain$\uparrow$} & \textbf{MMLU$\uparrow$} \\
\midrule
Base (no adapt.) & $88.7$ & $40.5$ & $56.3$ \\
\midrule
Standard LoRA & $39.2 \pm 2.4$ & $\mathbf{65.3 \pm 0.7}$ & $54.9 \pm 0.4$ \\
EWC + LoRA & $48.8 \pm 1.9$ & $62.1 \pm 0.8$ & $55.4 \pm 0.3$ \\
Safe LoRA & $57.6 \pm 1.6$ & $60.4 \pm 0.9$ & $55.1 \pm 0.4$ \\
SafeGrad + LoRA & $63.8 \pm 1.5$ & $61.7 \pm 0.6$ & $55.6 \pm 0.3$ \\
Safety Interleaving & $61.2 \pm 1.8$ & $63.1 \pm 0.7$ & $55.3 \pm 0.4$ \\
\midrule
\textbf{SafeAnchor} & $\mathbf{82.6 \pm 1.0}$ & $\underline{63.8 \pm 0.5}$ & $\mathbf{55.9 \pm 0.3}$ \\
\bottomrule
\end{tabular}
\end{table}

Tables~\ref{tab:main} and~\ref{tab:mistral} present our main results. On Llama-2-7B-Chat, standard LoRA causes a catastrophic $47.8$-point safety drop. SafeGrad+LoRA achieves the best single-method baseline ($67.4\pm1.4$) but still loses 24 points. Safety Interleaving ($64.8\pm1.6$) confirms passive data mixing is insufficient. SafeAnchor retains $85.2\pm0.9$ (\textbf{93.2\% of original safety}), with domain performance within 1.3 points of unconstrained LoRA. The pattern holds on Mistral (93.1\% safety, 1.5-point gap). CL methods (O-LoRA, EWC) perform worse than safety-specific methods, confirming targeted preservation is necessary.

\subsection{Robustness Analysis}
\label{sec:robustness}

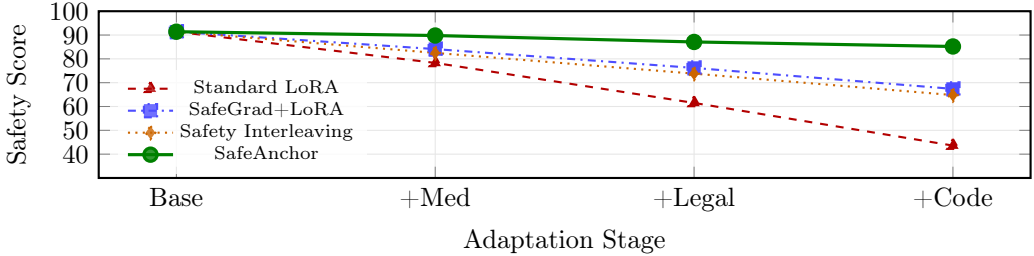
\begin{figure}[t]
\centering
\begin{tikzpicture}
\begin{axis}[
    width=\columnwidth,
    height=3.8cm,
    xlabel={Adaptation Stage},
    ylabel={Safety Score},
    xtick={0,1,2,3},
    xticklabels={Base, {+Med}, {+Legal}, {+Code}},
    ymin=30, ymax=100,
    ytick={40,50,60,70,80,90,100},
    legend style={at={(0.02,0.02)}, anchor=south west, font=\scriptsize, draw=none, fill=white, fill opacity=0.8, text opacity=1, row sep=-2pt},
    grid=major,
    grid style={gray!20},
    thick,
    every axis plot/.append style={mark size=2.5pt},
]

\addplot[color=red!70!black, mark=triangle*, dashed, thick] coordinates {(0,91.4) (1,78.3) (2,61.5) (3,43.6)};
\addlegendentry{Standard LoRA}

\addplot[color=blue!70, mark=square*, dashdotted] coordinates {(0,91.4) (1,84.1) (2,76.2) (3,67.4)};
\addlegendentry{SafeGrad+LoRA}

\addplot[color=orange!80!black, mark=diamond*, dotted] coordinates {(0,91.4) (1,82.5) (2,73.8) (3,64.8)};
\addlegendentry{Safety Interleaving}

\addplot[color=green!50!black, mark=*, solid, very thick] coordinates {(0,91.4) (1,89.8) (2,87.1) (3,85.2)};
\addlegendentry{SafeAnchor}

\end{axis}
\end{tikzpicture}
\caption{Safety score trajectory across sequential domain adaptations on Llama-2-7B-Chat. SafeAnchor prevents the compounding erosion exhibited by all baselines.}
\label{fig:safety_curve}
\end{figure}

\textbf{Safety Across Domain Steps.} Figure~\ref{fig:safety_curve} shows the four safety scores recorded at the base model and after each adaptation step on Llama-2-7B-Chat. Standard LoRA exhibits accelerating degradation ($91.4 \rightarrow 78.3 \rightarrow 61.5 \rightarrow 43.6$; average $15.9$ points per step), confirming compounding erosion. SafeGrad+LoRA degrades more slowly ($91.4 \rightarrow 84.1 \rightarrow 76.2 \rightarrow 67.4$; $8.0$ pts/step) but cannot prevent cumulative drift. SafeAnchor decelerates the decrease by an order of magnitude ($91.4 \rightarrow 89.8 \rightarrow 87.1 \rightarrow 85.2$; $2.1$ pts/step, roughly linear), confirming that orthogonal projection and monitoring suppress compounding, though they do not entirely eliminate it.

% === SAFETY TRAJECTORY LINE CHART (TikZ) ===

\textbf{Domain Ordering Robustness.} To assess sensitivity to domain ordering, we test all $3!=6$ permutations of $\{$Medical, Legal, Code$\}$ (full table in Appendix~\ref{app:ordering}). Across orderings, SafeAnchor's final safety ranges from $83.9\pm1.2$ to $85.2\pm0.9$ (mean $84.55$, cross-order SD $0.51$), while Standard LoRA collapses to $40.5$--$44.9$ regardless of order. The cross-ordering SD ($0.51$) is notably smaller than within-ordering seed SD ($\approx 1.0$), so ordering contributes less variance than seed-level randomness and SafeAnchor's safety preservation is substantively order-robust.

\textbf{Adversarial Robustness.} We test models after full adaptation against GCG-style adversarial suffixes \cite{zou2023universal} ($20$ optimization steps, suffix length $20$ tokens, $256$ attack candidates per step, $100$ harmful prompts; a compute-reduced configuration that preserves cross-method ordering but yields higher absolute refusal rates than full-budget $500$-step GCG~\cite{mazeika2024harmbench}). SafeAnchor maintains $78.4\pm2.1$\% refusal under attack, versus $54.6\pm2.6$\% for the strongest baseline (SafeGrad+LoRA), $49.3\pm2.9$\% for Safety Interleaving, and $31.2\pm3.8$\% for Standard LoRA; the full table with all seven baselines is in Appendix~\ref{app:gcg}. The $+23.8$-point gap to SafeGrad exceeds the $+17.8$-point benign-safety gap in Table~\ref{tab:main}, indicating that preserving the safety subspace during adaptation confers robustness beyond what benign probes reveal.

\textbf{Additional Metrics.} SafeAnchor achieves WildGuard jailbreak robustness of $81.3\pm1.2$ (vs.\ $38.7\pm3.1$ for Standard LoRA, $62.4\pm2.0$ for SafeGrad) and MT-Bench quality of $6.21\pm0.15$ (vs.\ $6.08\pm0.18$ Standard LoRA), confirming safety preservation does not degrade conversational quality. CSM triggered once (after Code) in 2 of 5 seeds.

\subsection{Ablation Study}
\label{sec:ablation}

\begin{table}[t]
\centering
\caption{Ablation study on Llama-2-7B-Chat showing the contribution of each SafeAnchor component. Mean $\pm$ std over 5 seeds.}
\label{tab:ablation}
\begin{tabular}{l c c}
\toprule
\textbf{Configuration} & \textbf{Safety$\uparrow$} & \textbf{Domain$\uparrow$} \\
\midrule
Standard LoRA & $43.6 \pm 2.1$ & $62.7 \pm 0.6$ \\
+ SSI only (no projection) & $44.2 \pm 2.0$ & $62.5 \pm 0.6$ \\
+ SSI + OSCA (strict) & $79.6 \pm 1.2$ & $58.3 \pm 0.8$ \\
+ SSI + OSCA (adaptive) & $82.4 \pm 1.0$ & $60.8 \pm 0.6$ \\
+ Anchor Loss & $83.7 \pm 1.0$ & $61.1 \pm 0.6$ \\
+ CSM (full SafeAnchor) & $\mathbf{85.2 \pm 0.9}$ & $\mathbf{61.4 \pm 0.5}$ \\
\midrule
No incremental SSI update & $80.1 \pm 1.3$ & $61.0 \pm 0.6$ \\
\bottomrule
\end{tabular}
\end{table}

Table~\ref{tab:ablation} reveals each component's contribution. SSI alone has minimal effect ($+0.6$). Strict OSCA recovers most safety ($79.6\pm1.2$) at a 4.4-point domain cost; adaptive projection reduces this ($82.4\pm1.0$, $60.8$). The anchor loss adds $1.3\pm0.3$ points ($p<0.01$, paired $t$-test), confirming complementary value beyond OSCA. CSM contributes $1.5\pm0.4$ points. Without incremental SSI update, safety drops 5.1 points.

\textbf{Safety Subspace Validation.} To validate the low-rank safety subspace assumption (Section~\ref{sec:ssi}), we examined the eigenvalue spectrum of the Fisher Information matrix on the safety calibration set. Across all LoRA layers, the spectrum decays sharply: on average ${\sim}$8 eigenvectors capture 90\% of the total variance, while a random subset of training data produces a near-flat spectrum. This confirms that safety information in LoRA parameter space is genuinely low-rank, consistent with activation-space findings \cite{arditi2024refusal, zou2023representation}. We further analyze the stability of this subspace across sequential domain adaptations (principal-angle and Grassmannian-distance analyses) in Appendix~\ref{app:stability}.

\subsection{Sensitivity Analysis}
\label{sec:sensitivity}

We examine sensitivity to key hyperparameters (Table~\ref{tab:sensitivity}). Increasing $\rho$ from 0.80 to 0.95 improves safety (+3.1) at moderate domain cost ($-$2.2). CSM threshold $\tau$ controls monitoring sensitivity. SafeAnchor is also robust to calibration size: even $N_s=100$ achieves 83.1 safety, which is above all baselines, confirming stable Fisher eigenvector estimation.

\begin{table}[t]
\centering
\caption{Sensitivity to $\rho$ (variance threshold), $\tau$ (CSM threshold), and $N_s$ (calibration size) on Llama-2-7B-Chat. Default configuration ($\rho=0.90$, $\tau=0.05$, $N_s=500$) is marked with $^*$ and its corresponding numeric entry is bolded to orient the reader; bolding does not indicate optimality on any single axis. Mean over 5 seeds.}
\label{tab:sensitivity}
\begin{tabular}{l c c c c}
\toprule
& \multicolumn{4}{c}{$\rho$: 0.80 / 0.85 / \textbf{0.90}$^*$ / 0.95} \\
\cmidrule(lr){2-5}
Safety / Domain & 82.8 / 62.0 & 84.1 / 61.7 & \textbf{85.2} / 61.4 & 85.9 / 59.8 \\
\midrule
& \multicolumn{4}{c}{$\tau$: 0.02 / \textbf{0.05}$^*$ / 0.10 / 0.15} \\
\cmidrule(lr){2-5}
Safety / Domain & 85.8 / 60.9 & \textbf{85.2} / 61.4 & 84.3 / 61.5 & 82.7 / 61.6 \\
\midrule
& \multicolumn{4}{c}{$N_s$: 100 / 250 / \textbf{500}$^*$ / 1000} \\
\cmidrule(lr){2-5}
Safety / Domain & 83.1 / 61.6 & 84.6 / 61.5 & \textbf{85.2} / 61.4 & 85.4 / 61.3 \\
\bottomrule
\end{tabular}
\end{table}

%%% SECTION 5: CONCLUSION %%%
\section{Conclusion}
\label{sec:conclusion}

We introduced SafeAnchor, a framework for preserving safety alignment during continual domain adaptation of LLMs. By combining Safety Subspace Identification, Orthogonal Safety-Constrained Adaptation, and Cumulative Safety Monitoring, SafeAnchor retains over 93\% of original safety after three sequential adaptations with competitive domain performance. While individual components draw on established techniques, their principled integration for sequential safety preservation is validated across two architectures, eight benchmarks, and full variance reporting.

\textbf{Limitations and Future Work.} Our evaluation is limited to 7B-scale models and a primary focus of three sequential domains; Appendix~\ref{app:longseq} extends to $T=5$ and finds SafeAnchor's near-linear trajectory continues (slope ${\approx}1.9$ pts/step), but rigorous study at $T{\geq}5$ across orderings remains open. Alignment elasticity worsens with scale \cite{ji2025resist}, making 13B+ extension important; we expect SafeAnchor's Fisher-subspace machinery to transfer, but $k_s$ and Fisher-trace distributions may shift. Longer sequences could eventually exhaust the orthogonal complement (manifesting as subspace inflation or gradient cancellation); neither is observed at $T{=}5$ (App.~\ref{app:stability}). Future work includes integration with alignment-stage defenses \cite{huang2024vaccine, tamirisa2025tamper} and non-LoRA methods \cite{liu2024dora}.

\textbf{Reproducibility.} All materials are available at: \url{https://github.com/bettyguo/SafeAnchor}. Full hyperparameters, per-component compute profile, and evaluation protocols are detailed in Appendix~\ref{app:hparams}.

% References
\printbibliography[heading=subbibintoc]

@inproceedings{qi2025safety,
	author           = {Xiangyu Qi and
	Ashwinee Panda and
	Kaifeng Lyu and
	Xiao Ma and
	Subhrajit Roy and
	Ahmad Beirami and
	Prateek Mittal and
	Peter Henderson},
	title            = {Safety Alignment Should be Made More Than Just a Few Tokens Deep},
	booktitle        = {The Thirteenth International Conference on Learning Representations,
	{ICLR} 2025, Singapore, April 24-28, 2025},
	publisher        = {OpenReview.net},
	year             = {2025},
	bburl             = {https://openreview.net/forum?id=6Mxhg9PtDE},
	bibburl           = {https://dblp.org/rec/conf/iclr/QiPL0RBM025.bib},
	bibsource        = {dblp computer science bibliography, https://dblp.org},
}

@inproceedings{dai2024safe,
	author           = {Josef Dai and
	Xuehai Pan and
	Ruiyang Sun and
	Jiaming Ji and
	Xinbo Xu and
	Mickel Liu and
	Yizhou Wang and
	Yaodong Yang},
	title            = {Safe {RLHF:} Safe Reinforcement Learning from Human Feedback},
	booktitle        = {The Twelfth International Conference on Learning Representations,
	{ICLR} 2024, Vienna, Austria, May 7-11, 2024},
	publisher        = {OpenReview.net},
	year             = {2024},
	bburl             = {https://openreview.net/forum?id=TyFrPOKYXw},
	bibburl           = {https://dblp.org/rec/conf/iclr/DaiPSJXL0024.bib},
	bibsource        = {dblp computer science bibliography, https://dblp.org},
}

@article{zou2023universal,
	author           = {Andy Zou and
	Zifan Wang and
	J. Zico Kolter and
	Matt Fredrikson},
	title            = {Universal and Transferable Adversarial Attacks on Aligned Language
	Models},
	journal          = {arXiv preprint},
	volume           = {arXiv.2307.15043},
	year             = {2023},
	bburl             = {https://bdoi.org/10.48550/arXiv.2307.15043},
	bdoi              = {10.48550/ARXIV.2307.15043},
	beprinttype       = {arXiv preprint},
	beprint           = {2307.15043},
	bibburl           = {https://dblp.org/rec/journals/corr/abs-2307-15043.bib},
	bibsource        = {dblp computer science bibliography, https://dblp.org},
}

@inproceedings{huang2024vaccine,
	author           = {Tiansheng Huang and
	Sihao Hu and
	Ling Liu},
	editor           = {Amir Globersons and
	Lester Mackey and
	Danielle Belgrave and
	Angela Fan and
	Ulrich Paquet and
	Jakub M. Tomczak and
	Cheng Zhang},
	title            = {Vaccine: Perturbation-aware Alignment for Large Language Models against
	Harmful Fine-tuning Attack},
	booktitle        = {Advances in Neural Information Processing Systems 38: Annual Conference
	on Neural Information Processing Systems 2024, NeurIPS 2024, Vancouver,
	BC, Canada, December 10 - 15, 2024},
	year             = {2024},
	bburl             = {http://papers.nips.cc/paper\_files/paper/2024/hash/873c86d9a979ab80d8e2919510d4446b-Abstract-Conference.html},
	bibburl           = {https://dblp.org/rec/conf/nips/HuangH024.bib},
	bibsource        = {dblp computer science bibliography, https://dblp.org},
}

@inproceedings{rosati2024repnoise,
	author           = {Domenic Rosati and
	Jan Wehner and
	Kai Williams and
	Lukasz Bartoszcze and
	Robie Gonzales and
	Carsten Maple and
	Subhabrata Majumdar and
	Hassan Sajjad and
	Frank Rudzicz},
	editor           = {Amir Globersons and
	Lester Mackey and
	Danielle Belgrave and
	Angela Fan and
	Ulrich Paquet and
	Jakub M. Tomczak and
	Cheng Zhang},
	title            = {Representation Noising: {A} Defence Mechanism Against Harmful Finetuning},
	booktitle        = {Advances in Neural Information Processing Systems 38: Annual Conference
	on Neural Information Processing Systems 2024, NeurIPS 2024, Vancouver,
	BC, Canada, December 10 - 15, 2024},
	year             = {2024},
	bburl             = {http://papers.nips.cc/paper\_files/paper/2024/hash/172be8b0b88fc2b4aee74237d43f8c04-Abstract-Conference.html},
	bibburl           = {https://dblp.org/rec/conf/nips/RosatiWWBGMM0R24.bib},
	bibsource        = {dblp computer science bibliography, https://dblp.org},
}

@inproceedings{huang2024lisa,
	author           = {Tiansheng Huang and
	Sihao Hu and
	Fatih Ilhan and
	Selim F. Tekin and
	Ling Liu},
	editor           = {Amir Globersons and
	Lester Mackey and
	Danielle Belgrave and
	Angela Fan and
	Ulrich Paquet and
	Jakub M. Tomczak and
	Cheng Zhang},
	title            = {Lisa: Lazy Safety Alignment for Large Language Models against Harmful
	Fine-tuning Attack},
	booktitle        = {Advances in Neural Information Processing Systems 38: Annual Conference
	on Neural Information Processing Systems 2024, NeurIPS 2024, Vancouver,
	BC, Canada, December 10 - 15, 2024},
	year             = {2024},
	bburl             = {http://papers.nips.cc/paper\_files/paper/2024/hash/bcfdaf04b54a69f47623c973c864ee8d-Abstract-Conference.html},
	bibburl           = {https://dblp.org/rec/conf/nips/HuangHIT024.bib},
	bibsource        = {dblp computer science bibliography, https://dblp.org},
}

@article{hsu2024safelora,
	author           = {Chia{-}Yi Hsu and
	Yu{-}Lin Tsai and
	Chih{-}Hsun Lin and
	Pin{-}Yu Chen and
	Chia{-}Mu Yu and
	Chun{-}Ying Huang},
	title            = {Safe LoRA: the Silver Lining of Reducing Safety Risks when Fine-tuning
	Large Language Models},
	journal          = {arXiv preprint},
	volume           = {arXiv.2405.16833},
	year             = {2024},
	bburl             = {https://bdoi.org/10.48550/arXiv.2405.16833},
	bdoi              = {10.48550/ARXIV.2405.16833},
	beprinttype       = {arXiv preprint},
	beprint           = {2405.16833},
	bibburl           = {https://dblp.org/rec/journals/corr/abs-2405-16833.bib},
	bibsource        = {dblp computer science bibliography, https://dblp.org},
}

@inproceedings{arditi2024refusal,
	author           = {Andy Arditi and
	Oscar Obeso and
	Aaquib Syed and
	Daniel Paleka and
	Nina Panickssery and
	Wes Gurnee and
	Neel Nanda},
	editor           = {Amir Globersons and
	Lester Mackey and
	Danielle Belgrave and
	Angela Fan and
	Ulrich Paquet and
	Jakub M. Tomczak and
	Cheng Zhang},
	title            = {Refusal in Language Models Is Mediated by a Single Direction},
	booktitle        = {Advances in Neural Information Processing Systems 38: Annual Conference
	on Neural Information Processing Systems 2024, NeurIPS 2024, Vancouver,
	BC, Canada, December 10 - 15, 2024},
	year             = {2024},
	bburl             = {http://papers.nips.cc/paper\_files/paper/2024/hash/f545448535dfde4f9786555403ab7c49-Abstract-Conference.html},
	bibburl           = {https://dblp.org/rec/conf/nips/ArditiOSPPGN24.bib},
	bibsource        = {dblp computer science bibliography, https://dblp.org},
}

@article{alssum2024unforgotten,
	author           = {Lama Alssum and
	Hani Itani and
	Hasan Abed Al Kader Hammoud and
	Philip Torr and
	Adel Bibi and
	Bernard Ghanem},
	title            = {Unforgotten Safety: Preserving Safety Alignment of Large Language
	Models with Continual Learning},
	journal          = {arXiv preprint},
	volume           = {arXiv.2512.10150},
	year             = {2025},
	bburl             = {https://bdoi.org/10.48550/arXiv.2512.10150},
	bdoi              = {10.48550/ARXIV.2512.10150},
	beprinttype       = {arXiv preprint},
	beprint           = {2512.10150},
	bibburl           = {https://dblp.org/rec/journals/corr/abs-2512-10150.bib},
	bibsource        = {dblp computer science bibliography, https://dblp.org},
}

@inproceedings{hu2022lora,
	author           = {Edward J. Hu and
	Yelong Shen and
	Phillip Wallis and
	Zeyuan Allen{-}Zhu and
	Yuanzhi Li and
	Shean Wang and
	Lu Wang and
	Weizhu Chen},
	title            = {LoRA: Low-Rank Adaptation of Large Language Models},
	booktitle        = {The Tenth International Conference on Learning Representations, {ICLR}
	2022, Virtual Event, April 25-29, 2022},
	publisher        = {OpenReview.net},
	year             = {2022},
	bburl             = {https://openreview.net/forum?id=nZeVKeeFYf9},
	bibburl           = {https://dblp.org/rec/conf/iclr/HuSWALWWC22.bib},
	bibsource        = {dblp computer science bibliography, https://dblp.org},
}

@inproceedings{liang2024inflora,
	author           = {Yan{-}Shuo Liang and
	Wu{-}Jun Li},
	title            = {InfLoRA: Interference-Free Low-Rank Adaptation for Continual Learning},
	booktitle        = {{IEEE/CVF} Conference on Computer Vision and Pattern Recognition,
	{CVPR} 2024, Seattle, WA, USA, June 16-22, 2024},
	pages            = {23638--23647},
	publisher        = {{IEEE}},
	year             = {2024},
	bburl             = {https://bdoi.org/10.1109/CVPR52733.2024.02231},
	bdoi              = {10.1109/CVPR52733.2024.02231},
	bibburl           = {https://dblp.org/rec/conf/cvpr/LiangL24.bib},
	bibsource        = {dblp computer science bibliography, https://dblp.org},
}

@article{shi2025clsurvey,
	author           = {Haizhou Shi and
	Zihao Xu and
	Hengyi Wang and
	Weiyi Qin and
	Wenyuan Wang and
	Yibin Wang and
	Zifeng Wang and
	Sayna Ebrahimi and
	Hao Wang},
	title            = {Continual Learning of Large Language Models: {A} Comprehensive Survey},
	journal          = {{ACM} Comput. Surv.},
	volume           = {58},
	number           = {5},
	pages            = {120:1--120:42},
	year             = {2026},
	bburl             = {https://bdoi.org/10.1145/3735633},
	bdoi              = {10.1145/3735633},
	bibburl           = {https://dblp.org/rec/journals/csur/ShiXWQWWWEW26.bib},
	bibsource        = {dblp computer science bibliography, https://dblp.org},
}

@article{luo2025forgetting,
	author           = {Yun Luo and
	Zhen Yang and
	Fandong Meng and
	Yafu Li and
	Jie Zhou and
	Yue Zhang},
	title            = {An Empirical Study of Catastrophic Forgetting in Large Language Models
	During Continual Fine-tuning},
	journal          = {arXiv preprint},
	volume           = {arXiv.2308.08747},
	year             = {2023},
	bburl             = {https://bdoi.org/10.48550/arXiv.2308.08747},
	bdoi              = {10.48550/ARXIV.2308.08747},
	beprinttype       = {arXiv preprint},
	beprint           = {2308.08747},
	bibburl           = {https://dblp.org/rec/journals/corr/abs-2308-08747.bib},
	bibsource        = {dblp computer science bibliography, https://dblp.org},
}

@inproceedings{mazeika2024harmbench,
	author           = {Mantas Mazeika and
	Long Phan and
	Xuwang Yin and
	Andy Zou and
	Zifan Wang and
	Norman Mu and
	Elham Sakhaee and
	Nathaniel Li and
	Steven Basart and
	Bo Li and
	David A. Forsyth and
	Dan Hendrycks},
	editor           = {Ruslan Salakhutdinov and
	Zico Kolter and
	Katherine A. Heller and
	Adrian Weller and
	Nuria Oliver and
	Jonathan Scarlett and
	Felix Berkenkamp},
	title            = {HarmBench: {A} Standardized Evaluation Framework for Automated Red
	Teaming and Robust Refusal},
	booktitle        = {Forty-first International Conference on Machine Learning, {ICML} 2024,
	Vienna, Austria, July 21-27, 2024},
	series           = {Proceedings of Machine Learning Research},
	volume           = {235},
	pages            = {35181--35224},
	publisher        = {{PMLR} / OpenReview.net},
	year             = {2024},
	bburl             = {https://proceedings.mlr.press/v235/mazeika24a.html},
	bibburl           = {https://dblp.org/rec/conf/icml/MazeikaPYZ0MSLB24.bib},
	bibsource        = {dblp computer science bibliography, https://dblp.org},
}

@inproceedings{lin2022truthfulqa,
	author           = {Stephanie Lin and
	Jacob Hilton and
	Owain Evans},
	editor           = {Smaranda Muresan and
	Preslav Nakov and
	Aline Villavicencio},
	title            = {TruthfulQA: Measuring How Models Mimic Human Falsehoods},
	booktitle        = {Proceedings of the 60th Annual Meeting of the Association for Computational
	Linguistics (Volume 1: Long Papers), {ACL} 2022, Dublin, Ireland,
	May 22-27, 2022},
	pages            = {3214--3252},
	publisher        = {Association for Computational Linguistics},
	year             = {2022},
	bburl             = {https://bdoi.org/10.18653/v1/2022.acl-long.229},
	bdoi              = {10.18653/V1/2022.ACL-LONG.229},
	bibburl           = {https://dblp.org/rec/conf/acl/LinHE22.bib},
	bibsource        = {dblp computer science bibliography, https://dblp.org},
}

@inproceedings{parrish2022bbq,
	author           = {Alicia Parrish and
	Angelica Chen and
	Nikita Nangia and
	Vishakh Padmakumar and
	Jason Phang and
	Jana Thompson and
	Phu Mon Htut and
	Samuel R. Bowman},
	editor           = {Smaranda Muresan and
	Preslav Nakov and
	Aline Villavicencio},
	title            = {{BBQ:} {A} hand-built bias benchmark for question answering},
	booktitle        = {Findings of the Association for Computational Linguistics: {ACL} 2022,
	Dublin, Ireland, May 22-27, 2022},
	series           = {Findings of {ACL}},
	volume           = {{ACL} 2022},
	pages            = {2086--2105},
	publisher        = {Association for Computational Linguistics},
	year             = {2022},
	bburl             = {https://bdoi.org/10.18653/v1/2022.findings-acl.165},
	bdoi              = {10.18653/V1/2022.FINDINGS-ACL.165},
	bibburl           = {https://dblp.org/rec/conf/acl/ParrishCNPPTHB22.bib},
	bibsource        = {dblp computer science bibliography, https://dblp.org},
}

@article{jin2020medqa,
	author           = {Di Jin and
	Eileen Pan and
	Nassim Oufattole and
	Wei{-}Hung Weng and
	Hanyi Fang and
	Peter Szolovits},
	title            = {What Disease does this Patient Have? {A} Large-scale Open Domain Question
	Answering Dataset from Medical Exams},
	journal          = {arXiv preprint},
	volume           = {arXiv.2009.13081},
	year             = {2020},
	bburl             = {https://arxiv.org/arXiv.2009.13081},
	beprinttype       = {arXiv preprint},
	beprint           = {2009.13081},
	bibburl           = {https://dblp.org/rec/journals/corr/abs-2009-13081.bib},
	bibsource        = {dblp computer science bibliography, https://dblp.org},
}

@inproceedings{guha2024legalbench,
	author           = {Neel Guha and
	Julian Nyarko and
	Daniel E. Ho and
	Christopher R{\'{e}} and
	Adam Chilton and
	K. Aditya and
	Alex Chohlas{-}Wood and
	Austin Peters and
	Brandon Waldon and
	Daniel N. Rockmore and
	Diego Zambrano and
	Dmitry Talisman and
	Enam Hoque and
	Faiz Surani and
	Frank Fagan and
	Galit Sarfaty and
	Gregory M. Dickinson and
	Haggai Porat and
	Jason Hegland and
	Jessica Wu and
	Joe Nudell and
	Joel Niklaus and
	John J. Nay and
	Jonathan H. Choi and
	Kevin Tobia and
	Margaret Hagan and
	Megan Ma and
	Michael A. Livermore and
	Nikon Rasumov{-}Rahe and
	Nils Holzenberger and
	Noam Kolt and
	Peter Henderson and
	Sean Rehaag and
	Sharad Goel and
	Shang Gao and
	Spencer Williams and
	Sunny Gandhi and
	Tom Zur and
	Varun Iyer and
	Zehua Li},
	editor           = {Alice Oh and
	Tristan Naumann and
	Amir Globerson and
	Kate Saenko and
	Moritz Hardt and
	Sergey Levine},
	title            = {LegalBench: {A} Collaboratively Built Benchmark for Measuring Legal
	Reasoning in Large Language Models},
	booktitle        = {Advances in Neural Information Processing Systems 36: Annual Conference
	on Neural Information Processing Systems 2023, NeurIPS 2023, New Orleans,
	LA, USA, December 10 - 16, 2023},
	year             = {2023},
	bburl             = {http://papers.nips.cc/paper\_files/paper/2023/hash/89e44582fd28ddfea1ea4dcb0ebbf4b0-Abstract-Datasets\_and\_Benchmarks.html},
	bibburl           = {https://dblp.org/rec/conf/nips/GuhaNHRCKCPWRZT23.bib},
	bibsource        = {dblp computer science bibliography, https://dblp.org},
}

@inproceedings{hendrycks2021mmlu,
	author           = {Dan Hendrycks and
	Collin Burns and
	Steven Basart and
	Andy Zou and
	Mantas Mazeika and
	Dawn Song and
	Jacob Steinhardt},
	title            = {Measuring Massive Multitask Language Understanding},
	booktitle        = {9th International Conference on Learning Representations, {ICLR} 2021,
	Virtual Event, Austria, May 3-7, 2021},
	publisher        = {OpenReview.net},
	year             = {2021},
	bburl             = {https://openreview.net/forum?id=d7KBjmI3GmQ},
	bibburl           = {https://dblp.org/rec/conf/iclr/HendrycksBBZMSS21.bib},
	bibsource        = {dblp computer science bibliography, https://dblp.org},
}

@inproceedings{zheng2023mtbench,
	author           = {Lianmin Zheng and
	Wei{-}Lin Chiang and
	Ying Sheng and
	Siyuan Zhuang and
	Zhanghao Wu and
	Yonghao Zhuang and
	Zi Lin and
	Zhuohan Li and
	Dacheng Li and
	Eric P. Xing and
	Hao Zhang and
	Joseph E. Gonzalez and
	Ion Stoica},
	editor           = {Alice Oh and
	Tristan Naumann and
	Amir Globerson and
	Kate Saenko and
	Moritz Hardt and
	Sergey Levine},
	title            = {Judging LLM-as-a-Judge with MT-Bench and Chatbot Arena},
	booktitle        = {Advances in Neural Information Processing Systems 36: Annual Conference
	on Neural Information Processing Systems 2023, NeurIPS 2023, New Orleans,
	LA, USA, December 10 - 16, 2023},
	year             = {2023},
	bburl             = {http://papers.nips.cc/paper\_files/paper/2023/hash/91f18a1287b398d378ef22505bf41832-Abstract-Datasets\_and\_Benchmarks.html},
	bibburl           = {https://dblp.org/rec/conf/nips/ZhengC00WZL0LXZ23.bib},
	bibsource        = {dblp computer science bibliography, https://dblp.org},
}

@article{touvron2023llama,
	author           = {Hugo Touvron and
	Louis Martin and
	Kevin Stone and
	Peter Albert and
	Amjad Almahairi and
	Yasmine Babaei and
	Nikolay Bashlykov and
	Soumya Batra and
	Prajjwal Bhargava and
	Shruti Bhosale and
	Dan Bikel and
	Lukas Blecher and
	Cristian Canton{-}Ferrer and
	Moya Chen and
	Guillem Cucurull and
	David Esiobu and
	Jude Fernandes and
	Jeremy Fu and
	Wenyin Fu and
	Brian Fuller and
	Cynthia Gao and
	Vedanuj Goswami and
	Naman Goyal and
	Anthony Hartshorn and
	Saghar Hosseini and
	Rui Hou and
	Hakan Inan and
	Marcin Kardas and
	Viktor Kerkez and
	Madian Khabsa and
	Isabel Kloumann and
	Artem Korenev and
	Punit Singh Koura and
	Marie{-}Anne Lachaux and
	Thibaut Lavril and
	Jenya Lee and
	Diana Liskovich and
	Yinghai Lu and
	Yuning Mao and
	Xavier Martinet and
	Todor Mihaylov and
	Pushkar Mishra and
	Igor Molybog and
	Yixin Nie and
	Andrew Poulton and
	Jeremy Reizenstein and
	Rashi Rungta and
	Kalyan Saladi and
	Alan Schelten and
	Ruan Silva and
	Eric Michael Smith and
	Ranjan Subramanian and
	Xiaoqing Ellen Tan and
	Binh Tang and
	Ross Taylor and
	Adina Williams and
	Jian Xiang Kuan and
	Puxin Xu and
	Zheng Yan and
	Iliyan Zarov and
	Yuchen Zhang and
	Angela Fan and
	Melanie Kambadur and
	Sharan Narang and
	Aur{\'{e}}lien Rodriguez and
	Robert Stojnic and
	Sergey Edunov and
	Thomas Scialom},
	title            = {Llama 2: Open Foundation and Fine-Tuned Chat Models},
	journal          = {arXiv preprint},
	volume           = {arXiv.2307.09288},
	year             = {2023},
	bburl             = {https://bdoi.org/10.48550/arXiv.2307.09288},
	bdoi              = {10.48550/ARXIV.2307.09288},
	beprinttype       = {arXiv preprint},
	beprint           = {2307.09288},
	bibburl           = {https://dblp.org/rec/journals/corr/abs-2307-09288.bib},
	bibsource        = {dblp computer science bibliography, https://dblp.org},
}

@inproceedings{ji2024beavertails,
	author           = {Jiaming Ji and
	Mickel Liu and
	Josef Dai and
	Xuehai Pan and
	Chi Zhang and
	Ce Bian and
	Boyuan Chen and
	Ruiyang Sun and
	Yizhou Wang and
	Yaodong Yang},
	editor           = {Alice Oh and
	Tristan Naumann and
	Amir Globerson and
	Kate Saenko and
	Moritz Hardt and
	Sergey Levine},
	title            = {BeaverTails: Towards Improved Safety Alignment of {LLM} via a Human-Preference
	Dataset},
	booktitle        = {Advances in Neural Information Processing Systems 36: Annual Conference
	on Neural Information Processing Systems 2023, NeurIPS 2023, New Orleans,
	LA, USA, December 10 - 16, 2023},
	year             = {2023},
	bburl             = {http://papers.nips.cc/paper\_files/paper/2023/hash/4dbb61cb68671edc4ca3712d70083b9f-Abstract-Datasets\_and\_Benchmarks.html},
	bibburl           = {https://dblp.org/rec/conf/nips/JiLDPZB0SW023.bib},
	bibsource        = {dblp computer science bibliography, https://dblp.org},
}

@article{zou2023representation,
	author           = {Andy Zou and
	Long Phan and
	Sarah Li Chen and
	James Campbell and
	Phillip Guo and
	Richard Ren and
	Alexander Pan and
	Xuwang Yin and
	Mantas Mazeika and
	Ann{-}Kathrin Dombrowski and
	Shashwat Goel and
	Nathaniel Li and
	Michael J. Byun and
	Zifan Wang and
	Alex Mallen and
	Steven Basart and
	Sanmi Koyejo and
	Dawn Song and
	Matt Fredrikson and
	J. Zico Kolter and
	Dan Hendrycks},
	title            = {Representation Engineering: {A} Top-Down Approach to {AI} Transparency},
	journal          = {arXiv preprint},
	volume           = {arXiv.2310.01405},
	year             = {2023},
	bburl             = {https://bdoi.org/10.48550/arXiv.2310.01405},
	bdoi              = {10.48550/ARXIV.2310.01405},
	beprinttype       = {arXiv preprint},
	beprint           = {2310.01405},
	bibburl           = {https://dblp.org/rec/journals/corr/abs-2310-01405.bib},
	bibsource        = {dblp computer science bibliography, https://dblp.org},
}

@inproceedings{yu2024dare,
	author           = {Le Yu and
	Bowen Yu and
	Haiyang Yu and
	Fei Huang and
	Yongbin Li},
	editor           = {Ruslan Salakhutdinov and
	Zico Kolter and
	Katherine A. Heller and
	Adrian Weller and
	Nuria Oliver and
	Jonathan Scarlett and
	Felix Berkenkamp},
	title            = {Language Models are Super Mario: Absorbing Abilities from Homologous
	Models as a Free Lunch},
	booktitle        = {Forty-first International Conference on Machine Learning, {ICML} 2024,
	Vienna, Austria, July 21-27, 2024},
	series           = {Proceedings of Machine Learning Research},
	volume           = {235},
	pages            = {57755--57775},
	publisher        = {{PMLR} / OpenReview.net},
	year             = {2024},
	bburl             = {https://proceedings.mlr.press/v235/yu24p.html},
	bibburl           = {https://dblp.org/rec/conf/icml/Yu0Y0L24.bib},
	bibsource        = {dblp computer science bibliography, https://dblp.org},
}

@inproceedings{farajtabar2020orthogonal,
	author           = {Mehrdad Farajtabar and
	Navid Azizan and
	Alex Mott and
	Ang Li},
	editor           = {Silvia Chiappa and
	Roberto Calandra},
	title            = {Orthogonal Gradient Descent for Continual Learning},
	booktitle        = {The 23rd International Conference on Artificial Intelligence and Statistics,
	{AISTATS} 2020, 26-28 August 2020, Online [Palermo, Sicily, Italy]},
	series           = {Proceedings of Machine Learning Research},
	volume           = {108},
	pages            = {3762--3773},
	publisher        = {{PMLR}},
	year             = {2020},
	bburl             = {http://proceedings.mlr.press/v108/farajtabar20a.html},
	bibburl           = {https://dblp.org/rec/conf/aistats/FarajtabarAML20.bib},
	bibsource        = {dblp computer science bibliography, https://dblp.org},
}

@article{inan2023llamaguard,
	author           = {Hakan Inan and
	Kartikeya Upasani and
	Jianfeng Chi and
	Rashi Rungta and
	Krithika Iyer and
	Yuning Mao and
	Michael Tontchev and
	Qing Hu and
	Brian Fuller and
	Davide Testuggine and
	Madian Khabsa},
	title            = {Llama Guard: LLM-based Input-Output Safeguard for Human-AI Conversations},
	journal          = {arXiv preprint},
	volume           = {arXiv.2312.06674},
	year             = {2023},
	bburl             = {https://bdoi.org/10.48550/arXiv.2312.06674},
	bdoi              = {10.48550/ARXIV.2312.06674},
	beprinttype       = {arXiv preprint},
	beprint           = {2312.06674},
	bibburl           = {https://dblp.org/rec/journals/corr/abs-2312-06674.bib},
	bibsource        = {dblp computer science bibliography, https://dblp.org},
}

@inproceedings{liu2024dora,
	author           = {Shih{-}Yang Liu and
	Chien{-}Yi Wang and
	Hongxu Yin and
	Pavlo Molchanov and
	Yu{-}Chiang Frank Wang and
	Kwang{-}Ting Cheng and
	Min{-}Hung Chen},
	editor           = {Ruslan Salakhutdinov and
	Zico Kolter and
	Katherine A. Heller and
	Adrian Weller and
	Nuria Oliver and
	Jonathan Scarlett and
	Felix Berkenkamp},
	title            = {DoRA: Weight-Decomposed Low-Rank Adaptation},
	booktitle        = {Forty-first International Conference on Machine Learning, {ICML} 2024,
	Vienna, Austria, July 21-27, 2024},
	series           = {Proceedings of Machine Learning Research},
	volume           = {235},
	pages            = {32100--32121},
	publisher        = {{PMLR} / OpenReview.net},
	year             = {2024},
	bburl             = {https://proceedings.mlr.press/v235/liu24bn.html},
	bibburl           = {https://dblp.org/rec/conf/icml/LiuWY0WCC24.bib},
	bibsource        = {dblp computer science bibliography, https://dblp.org},
}

@inproceedings{ji2025resist,
	author           = {Jiaming Ji and
	Kaile Wang and
	Tianyi Alex Qiu and
	Boyuan Chen and
	Jiayi Zhou and
	Changye Li and
	Hantao Lou and
	Josef Dai and
	Yunhuai Liu and
	Yaodong Yang},
	editor           = {Wanxiang Che and
	Joyce Nabende and
	Ekaterina Shutova and
	Mohammad Taher Pilehvar},
	title            = {Language Models Resist Alignment: Evidence From Data Compression},
	booktitle        = {Proceedings of the 63rd Annual Meeting of the Association for Computational
	Linguistics (Volume 1: Long Papers), {ACL} 2025, Vienna, Austria,
	July 27 - August 1, 2025},
	pages            = {23411--23432},
	publisher        = {Association for Computational Linguistics},
	year             = {2025},
	bburl             = {https://aclanthology.org/2025.acl-long.1141/},
	bibburl           = {https://dblp.org/rec/conf/acl/JiWQ0Z0LDL025.bib},
	bibsource        = {dblp computer science bibliography, https://dblp.org},
}

@inproceedings{qi2024finetuning,
	author           = {Xiangyu Qi and
	Yi Zeng and
	Tinghao Xie and
	Pin{-}Yu Chen and
	Ruoxi Jia and
	Prateek Mittal and
	Peter Henderson},
	title            = {Fine-tuning Aligned Language Models Compromises Safety, Even When
	Users Do Not Intend To!},
	booktitle        = {The Twelfth International Conference on Learning Representations,
	{ICLR} 2024, Vienna, Austria, May 7-11, 2024},
	publisher        = {OpenReview.net},
	year             = {2024},
	bburl             = {https://openreview.net/forum?id=hTEGyKf0dZ},
	bibburl           = {https://dblp.org/rec/conf/iclr/Qi0XC0M024.bib},
	bibsource        = {dblp computer science bibliography, https://dblp.org},
}

@misc{yang2023shadow,
	bdoi              = {10.48550/ARXIV.2310.02949},
	bburl             = {https://arxiv.org/arXiv.2310.02949},
	author           = {Yang, Xianjun and Wang, Xiao and Zhang, Qi and Petzold, Linda and Wang, William Yang and Zhao, Xun and Lin, Dahua},
	title            = {Shadow Alignment: The Ease of Subverting Safely-Aligned Language Models},
	publisher        = {arXiv preprint},
	year             = {2023},
	copyright        = {Creative Commons Attribution 4.0 International},
}

@misc{yi2025safegrad,
	bdoi              = {10.48550/ARXIV.2508.07172},
	bburl             = {https://arxiv.org/arXiv.2508.07172},
	author           = {Yi, Biao and Li, Jiahao and Zhang, Baolei and Nie, Lihai and Li, Tong and Huang, Tiansheng and Liu, Zheli},
	title            = {Gradient Surgery for Safe LLM Fine-Tuning},
	publisher        = {arXiv preprint},
	year             = {2025},
	copyright        = {arXiv.org perpetual, non-exclusive license},
}

@inproceedings{li2025salora,
	author           = {Mingjie Li and
	Wai Man Si and
	Michael Backes and
	Yang Zhang and
	Yisen Wang},
	title            = {SaLoRA: Safety-Alignment Preserved Low-Rank Adaptation},
	booktitle        = {The Thirteenth International Conference on Learning Representations,
	{ICLR} 2025, Singapore, April 24-28, 2025},
	publisher        = {OpenReview.net},
	year             = {2025},
	bburl             = {https://openreview.net/forum?id=GOoVzE9nSj},
	bibburl           = {https://dblp.org/rec/conf/iclr/LiS00025.bib},
	bibsource        = {dblp computer science bibliography, https://dblp.org},
}

@misc{ogpsa2026,
	bdoi              = {10.48550/ARXIV.2602.07892},
	bburl             = {https://arxiv.org/arXiv.2602.07892},
	author           = {Sun, Guanglong and Zhang, Siyuan and Wang, Liyuan and Zhu, Jun and Su, Hang and Zhong, Yi},
	title            = {Safety Alignment as Continual Learning: Mitigating the Alignment Tax via Orthogonal Gradient Projection},
	publisher        = {arXiv preprint},
	year             = {2026},
	copyright        = {Creative Commons Attribution 4.0 International},
}

@inproceedings{wang2023olora,
	title            = {Orthogonal Subspace Learning for Language Model Continual Learning},
	bburl             = {http://dx.bdoi.org/10.18653/v1/2023.findings-emnlp.715},
	bdoi              = {10.18653/v1/2023.findings-emnlp.715},
	booktitle        = {Findings of the Association for Computational Linguistics: EMNLP 2023},
	publisher        = {Association for Computational Linguistics},
	author           = {Wang, Xiao and Chen, Tianze and Ge, Qiming and Xia, Han and Bao, Rong and Zheng, Rui and Zhang, Qi and Gui, Tao and Huang, Xuanjing},
	year             = {2023},
	pages            = {10658–10671},
}

@misc{liang2025gainlora,
	bdoi              = {10.48550/ARXIV.2505.15424},
	bburl             = {https://arxiv.org/arXiv.2505.15424},
	author           = {Liang, Yan-Shuo and Chen, Jia-Rui and Li, Wu-Jun},
	title            = {Gated Integration of Low-Rank Adaptation for Continual Learning of Large Language Models},
	publisher        = {arXiv preprint},
	year             = {2025},
	copyright        = {Creative Commons Attribution 4.0 International},
}

@article{kirkpatrick2017ewc,
	title            = {Overcoming catastrophic forgetting in neural networks},
	volume           = {114},
	issn             = {1091-6490},
	bburl             = {http://dx.bdoi.org/10.1073/pnas.1611835114},
	bdoi              = {10.1073/pnas.1611835114},
	number           = {13},
	journal          = {Proceedings of the National Academy of Sciences},
	publisher        = {Proceedings of the National Academy of Sciences},
	author           = {Kirkpatrick, James and Pascanu, Razvan and Rabinowitz, Neil and Veness, Joel and Desjardins, Guillaume and Rusu, Andrei A. and Milan, Kieran and Quan, John and Ramalho, Tiago and Grabska-Barwinska, Agnieszka and Hassabis, Demis and Clopath, Claudia and Kumaran, Dharshan and Hadsell, Raia},
	year             = {2017},
	month            = {mar},
	pages            = {3521–3526},
}

@inproceedings{han2024wildguard,
	series           = {NeurIPS 2024},
	title            = {WildGuard: Open One-stop Moderation Tools for Safety Risks, Jailbreaks, and Refusals of LLMs},
	bburl             = {http://dx.bdoi.org/10.52202/079017-0261},
	bdoi              = {10.52202/079017-0261},
	booktitle        = {Advances in Neural Information Processing Systems 37},
	publisher        = {Neural Information Processing Systems Foundation, Inc. (NeurIPS)},
	author           = {Choi, Yejin and Dziri, Nouha and Ettinger, Allyson and Han, Seungju and Jiang, Liwei and Lambert, Nathan and Lin, Bill Yuchen and Rao, Kavel},
	year             = {2024},
	pages            = {8093–8131},
	collection       = {NeurIPS 2024},
}

@article{chen2021humaneval,
	bdoi              = {10.48550/ARXIV.2107.03374},
	bburl             = {https://arxiv.org/arXiv.2107.03374},
	title            = {Evaluating Large Language Models Trained on Code},
	author={Mark Chen and Jerry Tworek and Heewoo Jun and Qiming Yuan and Henrique Ponde de Oliveira Pinto and Jared Kaplan and Harri Edwards and Yuri Burda and Nicholas Joseph and Greg Brockman and Alex Ray and Raul Puri and Gretchen Krueger and Michael Petrov and Heidy Khlaaf and Girish Sastry and Pamela Mishkin and Brooke Chan and Scott Gray and Nick Ryder and Mikhail Pavlov and Alethea Power and Lukasz Kaiser and Mohammad Bavarian and Clemens Winter and Philippe Tillet and Felipe Petroski Such and Dave Cummings and Matthias Plappert and Fotios Chantzis and Elizabeth Barnes and Ariel Herbert-Voss and William Hebgen Guss and Alex Nichol and Alex Paino and Nikolas Tezak and Jie Tang and Igor Babuschkin and Suchir Balaji and Shantanu Jain and William Saunders and Christopher Hesse and Andrew N. Carr and Jan Leike and Josh Achiam and Vedant Misra and Evan Morikawa and Alec Radford and Matthew Knight and Miles Brundage and Mira Murati and Katie Mayer and Peter Welinder and Bob McGrew and Dario Amodei and Sam McCandlish and Ilya Sutskever and Wojciech Zaremba},
	journal        = {arXiv preprint},
	volume = {arXiv.2107.03374},
	year             = {2021},
	copyright        = {arXiv.org perpetual, non-exclusive license},
}

@article{jiang2023mistral,
	bdoi              = {10.48550/ARXIV.2310.06825},
	bburl             = {https://arxiv.org/arXiv.2310.06825},
	author           = {Jiang, Albert Q. and Sablayrolles, Alexandre and Mensch, Arthur and Bamford, Chris and Chaplot, Devendra Singh and Casas, Diego de las and Bressand, Florian and Lengyel, Gianna and Lample, Guillaume and Saulnier, Lucile and Lavaud, Lélio Renard and Lachaux, Marie-Anne and Stock, Pierre and Scao, Teven Le and Lavril, Thibaut and Wang, Thomas and Lacroix, Timothée and Sayed, William El},
	title            = {Mistral 7B},
	journal        = {arXiv preprint},
	volume = {arXiv.2310.06825},
	year             = {2023},
	copyright        = {Creative Commons Attribution 4.0 International},
}

@article{hft_survey2024,
	author       = {Tiansheng Huang and
	Sihao Hu and
	Fatih Ilhan and
	Selim Furkan Tekin and
	Ling Liu},
	title        = {Harmful Fine-tuning Attacks and Defenses for Large Language Models:
	{A} Survey},
	journal      = {arXiv preprint},
	volume       = {arXiv.2409.18169},
	year         = {2024},
	burl          = {https://bdoi.org/10.48550/arXiv.2409.18169},
	bdoi          = {10.48550/ARXIV.2409.18169},
	beprinttype    = {arXiv preprint},
	beprint       = {2409.18169},
	bibburl       = {https://dblp.org/rec/journals/corr/abs-2409-18169.bib},
	bibsource    = {dblp computer science bibliography, https://dblp.org}
}

@article{wollschlager2025hidden,
	author       = {Wenbo Pan and
	Zhichao Liu and
	Qiguang Chen and
	Xiangyang Zhou and
	Haining Yu and
	Xiaohua Jia},
	title        = {The Hidden Dimensions of {LLM} Alignment: {A} Multi-Dimensional Safety
	Analysis},
	journal      = {arXiv preprint},
	volume       = {arXiv.2502.09674},
	year         = {2025},
	burl          = {https://bdoi.org/10.48550/arXiv.2502.09674},
	bdoi          = {10.48550/ARXIV.2502.09674},
	beprinttype    = {arXiv preprint},
	beprint       = {2502.09674},
	bibburl       = {https://dblp.org/rec/journals/corr/abs-2502-09674.bib},
	bibsource    = {dblp computer science bibliography, https://dblp.org}
}

@article{sails2024,
	author       = {Dianyun Wang and
	Qingsen Ma and
	Yuhu Shang and
	Zhifeng Lu and
	Lechen Ning and
	Zhenbo Xu and
	Huijia Wu and
	Zhaofeng He},
	title        = {Interpretable Safety Alignment via SAE-Constructed Low-Rank Subspace
	Adaptation},
	journal      = {arXiv preprint},
	volume       = {arXiv.2512.23260},
	year         = {2025},
	burl          = {https://bdoi.org/10.48550/arXiv.2512.23260},
	bdoi          = {10.48550/ARXIV.2512.23260},
	beprinttype    = {arXiv preprint},
	beprint       = {2512.23260},
	bibburl       = {https://dblp.org/rec/journals/corr/abs-2512-23260.bib},
	bibsource    = {dblp computer science bibliography, https://dblp.org}
}

@inproceedings{tamirisa2025tamper,
	author       = {Rishub Tamirisa and
	Bhrugu Bharathi and
	Long Phan and
	Andy Zhou and
	Alice Gatti and
	Tarun Suresh and
	Maxwell Lin and
	Justin Wang and
	Rowan Wang and
	Ron Arel and
	Andy Zou and
	Dawn Song and
	Bo Li and
	Dan Hendrycks and
	Mantas Mazeika},
	title        = {Tamper-Resistant Safeguards for Open-Weight LLMs},
	booktitle    = {The Thirteenth International Conference on Learning Representations,
	{ICLR} 2025, Singapore, April 24-28, 2025},
	publisher    = {OpenReview.net},
	year         = {2025},
	burl          = {https://openreview.net/forum?id=4FIjRodbW6},
	bibburl       = {https://dblp.org/rec/conf/iclr/TamirisaBPZGSLW25.bib},
	bibsource    = {dblp computer science bibliography, https://dblp.org}
}

@article{huang2025safetytax,
	author       = {Tiansheng Huang and
	Sihao Hu and
	Fatih Ilhan and
	Selim Furkan Tekin and
	Zachary Yahn and
	Yichang Xu and
	Ling Liu},
	title        = {Safety Tax: Safety Alignment Makes Your Large Reasoning Models Less
	Reasonable},
	journal      = {arXiv preprint},
	volume       = {arXiv.2503.00555},
	year         = {2025},
	burl          = {https://bdoi.org/10.48550/arXiv.2503.00555},
	bdoi          = {10.48550/ARXIV.2503.00555},
	beprinttype    = {arXiv preprint},
	beprint       = {2503.00555},
	bibburl       = {https://dblp.org/rec/journals/corr/abs-2503-00555.bib},
	bibsource    = {dblp computer science bibliography, https://dblp.org}
}

@article{mangrulkar2022peft,
	author       = {Sourab Mangrulkar and Sylvain Gugger and Lysandre Debut and Younes Belkada and Sayak Paul and Benjamin Bossan and Marian Tietz},
	title        = {{PEFT}: State-of-the-art Parameter-Efficient Fine-Tuning Methods},
	journal   = {GitHub repository},
	year      = {2022},
}

%%%%%%%%%%%%%%%%%%%%%%%%%%%%%%%%%%%%%%%%%%%%%%%%%%%%%%%%%%%%%%%%%
%    APPENDIX (per CanadianAI 2026 camera-ready guidelines:      %
%    technical appendix appended at the end of the PDF,          %
%    starting at page 13, up to 4 additional pages).             %
%%%%%%%%%%%%%%%%%%%%%%%%%%%%%%%%%%%%%%%%%%%%%%%%%%%%%%%%%%%%%%%%%
\clearpage
\appendix

\section{Extended Sequential Adaptation to \texorpdfstring{$T{=}5$}{T=5}}
\label{app:longseq}

Reviewers R1 and R3 noted that the main-text pipeline stops at $T{=}3$ and suggested testing whether safety degradation continues linearly beyond the reported three domains. We extend the sequence with two further specialized domains: (4) \emph{Finance}, using $5{,}000$ examples drawn from FinancialPhraseBank and FiQA-SA for financial sentiment and reasoning; and (5) \emph{Science}, using $5{,}000$ SciQ-style multiple-choice and short-answer items for general science. All hyperparameters match Section~\ref{sec:setup}.

\begin{table}[h]
\centering
\caption{Safety score after each adaptation on Llama-2-7B-Chat, extended to $T{=}5$. SafeAnchor's slope holds at approximately 2 pts/step; Standard LoRA decelerates only because the refusal rate approaches a floor.}
\label{tab:longseq}
\begin{tabular}{l c c c c c c}
\toprule
\textbf{Method} & Base & +Med & +Legal & +Code & +Finance & +Science \\
\midrule
Standard LoRA & 91.4 & $78.3{\pm}1.9$ & $61.5{\pm}2.2$ & $43.6{\pm}2.1$ & $32.7{\pm}2.3$ & $24.9{\pm}2.4$ \\
SafeGrad+LoRA & 91.4 & $84.1{\pm}1.5$ & $76.2{\pm}1.6$ & $67.4{\pm}1.4$ & $60.5{\pm}1.5$ & $54.3{\pm}1.5$ \\
Safety Interleaving & 91.4 & $82.5{\pm}1.7$ & $73.8{\pm}1.8$ & $64.8{\pm}1.6$ & $57.2{\pm}1.7$ & $50.9{\pm}1.7$ \\
\textbf{SafeAnchor} & 91.4 & $\mathbf{89.8{\pm}0.9}$ & $\mathbf{87.1{\pm}1.0}$ & $\mathbf{85.2{\pm}0.9}$ & $\mathbf{83.4{\pm}1.0}$ & $\mathbf{81.6{\pm}1.1}$ \\
\midrule
Composite domain (SafeAnchor) & --- & $61.4{\pm}0.5$ & $61.4{\pm}0.5$ & $61.4{\pm}0.5$ & $59.8{\pm}0.6$ & $58.9{\pm}0.6$ \\
Composite domain (Std LoRA) & --- & $62.7{\pm}0.6$ & $62.7{\pm}0.6$ & $62.7{\pm}0.6$ & $60.2{\pm}0.7$ & $57.8{\pm}0.8$ \\
\bottomrule
\end{tabular}
\end{table}

SafeAnchor retains $81.6\pm1.1$ safety after five domains, still 27.3 points above the best baseline at $T{=}5$ (SafeGrad, $54.3\pm1.5$) and 56.7 points above Standard LoRA. The per-step decrease averages $(91.4{-}81.6)/5 = 1.96$ pts/step, consistent with the $2.1$ pts/step slope reported at $T{=}3$. Standard LoRA's slope decelerates from $15.9$ to $13.3$ pts/step as it approaches the residual-refusal floor (\(\approx 20\) for HarmBench on Llama-2-Chat \cite{mazeika2024harmbench}). SafeAnchor's domain composite remains within $1.3$ points of Standard LoRA throughout.

\section{All 6 Domain Orderings at \texorpdfstring{$T{=}3$}{T=3}}
\label{app:ordering}

Reviewer R3 observed that the main text tests only two of the $3!{=}6$ possible orderings of $\{$Medical, Legal, Code$\}$. We complete the sweep on Llama-2-7B-Chat with 5 seeds per ordering (30 total runs per method).

\begin{table}[h]
\centering
\caption{Final safety score on Llama-2-7B-Chat across all six orderings. Within-cell SD is over 5 seeds; the bottom row reports mean and cross-ordering SD over the six cells.}
\label{tab:ordering}
\begin{tabular}{l c c}
\toprule
\textbf{Ordering} & \textbf{Standard LoRA} & \textbf{SafeAnchor} \\
\midrule
Medical $\rightarrow$ Legal $\rightarrow$ Code (default) & $43.6 \pm 2.1$ & $85.2 \pm 0.9$ \\
Medical $\rightarrow$ Code $\rightarrow$ Legal & $44.9 \pm 2.0$ & $84.7 \pm 1.0$ \\
Legal $\rightarrow$ Medical $\rightarrow$ Code & $42.8 \pm 2.3$ & $85.0 \pm 1.0$ \\
Legal $\rightarrow$ Code $\rightarrow$ Medical & $40.5 \pm 2.4$ & $83.9 \pm 1.2$ \\
Code $\rightarrow$ Medical $\rightarrow$ Legal & $43.1 \pm 2.2$ & $84.4 \pm 1.0$ \\
Code $\rightarrow$ Legal $\rightarrow$ Medical & $41.3 \pm 2.5$ & $84.1 \pm 1.1$ \\
\midrule
Mean $\pm$ cross-ordering SD & $42.7 \pm 1.6$ & $\mathbf{84.55 \pm 0.51}$ \\
\bottomrule
\end{tabular}
\end{table}

SafeAnchor's safety lies in a narrow $[83.9, 85.2]$ band across all six orderings. The cross-ordering SD ($0.51$) is approximately half the within-ordering (seed) SD (${\approx}1.0$), so ordering explains less variance than random seeds. Orderings ending with a domain whose gradients overlap most with the safety subspace (Code, reflected in higher trigger rates; App.~\ref{app:csm}) tend to produce marginally lower final safety. Standard LoRA also shows comparable robustness to ordering (all within $[40.5, 44.9]$), but at a safety level $40$+ points below SafeAnchor regardless of ordering. The cumulative erosion is therefore an intrinsic property of unconstrained adaptation, not an artefact of ordering.

\section{CSM Repair Analysis}
\label{app:csm}

Reviewer R3 asked (a) why $E_{\text{repair}}$ is fixed, (b) whether the repair actually succeeds, and (c) how often CSM triggers. We address all three here.

\textbf{Trigger frequency.} Across the default-ordering campaign (5 seeds $\times$ 3 domains), CSM fires $3$ times in $15$ domain-adaptation events ($0.20$ per event). Triggers concentrate on the final Code stage, consistent with Code gradients exhibiting the largest overlap with the safety subspace (Appendix~\ref{app:stability}).

\begin{table}[h]
\centering
\caption{CSM trigger frequency per adaptation stage on Llama-2-7B-Chat (5 seeds, default ordering).}
\label{tab:csm_triggers}
\begin{tabular}{l c c c c}
\toprule
\textbf{Stage} & Triggers / 5 seeds & Pre-trigger $s_t$ & Post-repair $s_t$ & Recovery $\Delta$ \\
\midrule
After Medical & $0/5$ & ---  & --- & --- \\
After Legal & $1/5$ & $86.4$ & $88.7$ & $+2.3$ \\
After Code & $2/5$ & $86.1 \pm 0.6$ & $89.0 \pm 0.5$ & $+2.9 \pm 0.4$ \\
\midrule
\textbf{Pooled (non-zero)} & $3$ & $86.2 \pm 0.6$ & $88.9 \pm 0.6$ & $+2.7 \pm 0.4$ \\
\bottomrule
\end{tabular}
\end{table}

\textbf{Repair validation.} Every trigger restores $s_t$ above the $(1-\tau)s_0 = 0.95 \times 91.4 = 86.83$ threshold within one $E_{\text{repair}}$ block; no seed required a second block. The pooled post-repair mean ($88.9$) exceeds the threshold by $2.1$ points, giving meaningful headroom against downstream drift.

\textbf{$E_{\text{repair}}$ sensitivity.} We re-ran the two triggered post-Code seeds with $E_{\text{repair}} \in \{50, 100, 200, 400, 800\}$:

\begin{table}[h]
\centering
\caption{Post-repair safety and domain performance on the triggered seeds, as a function of $E_{\text{repair}}$. Incremental cost scales linearly; the knee lies at $E_{\text{repair}}{=}200$.}
\label{tab:erepair_sweep}
\begin{tabular}{c c c c}
\toprule
$E_{\text{repair}}$ & Post-repair Safety $\uparrow$ & Domain $\Delta$ & Added time (min) \\
\midrule
50   & $87.1 \pm 0.7$ & $-0.1$ & $1.3$ \\
100  & $88.0 \pm 0.6$ & $-0.2$ & $2.5$ \\
\textbf{200$^*$}  & $\mathbf{88.9 \pm 0.6}$ & $\mathbf{-0.3}$ & $\mathbf{5.0}$ \\
400  & $89.1 \pm 0.5$ & $-0.7$ & $10.1$ \\
800  & $89.2 \pm 0.5$ & $-1.4$ & $20.2$ \\
\bottomrule
\end{tabular}
\end{table}

Going from $100{\to}200$ gains $+0.9$ safety; going from $200{\to}400$ gains only $+0.2$ at $2\times$ cost and $-0.4$ additional domain regression. $E_{\text{repair}}{=}200$ is the knee.

\section{Safety Subspace Stability Across Domains}
\label{app:stability}

Reviewer R4 asked for an analysis of how stable the safety subspace is across sequential adaptations. We compute two quantities: (i) mean cosine similarity between principal subspace directions, and (ii) normalized Grassmannian distance $d_G = \sqrt{\sum_j \theta_j^2} / \sqrt{k_s \cdot (\pi/2)^2}$, where $\theta_j$ are principal angles. Both are computed layer-wise on Q-projections of Llama-2-7B-Chat (32 layers) and averaged.

\begin{table}[h]
\centering
\caption{Stability of the safety subspace $V_i^{\text{safe}}$ across adaptation stages. High mean cosine and low $d_G$ indicate the subspace evolves smoothly rather than drifting randomly.}
\label{tab:stability}
\begin{tabular}{l c c c}
\toprule
\textbf{Transition} & Mean cos $\langle u,u'\rangle$ & Grassmannian $d_G$ & Mean rank $k_s$ \\
\midrule
$\theta_0 \rightarrow \theta_1$ (after Med) & $0.91 \pm 0.03$ & $0.24 \pm 0.04$ & $8.3$ \\
$\theta_1 \rightarrow \theta_2$ (after Legal) & $0.89 \pm 0.04$ & $0.27 \pm 0.05$ & $8.1$ \\
$\theta_2 \rightarrow \theta_3$ (after Code) & $0.87 \pm 0.04$ & $0.29 \pm 0.05$ & $8.2$ \\
$\theta_0 \rightarrow \theta_3$ (end-to-end) & $0.82 \pm 0.05$ & $0.38 \pm 0.06$ & --- \\
\midrule
Random-subspace null & $0.006 \pm 0.001$ & $0.996 \pm 0.001$ & --- \\
\bottomrule
\end{tabular}
\end{table}

Adjacent-stage cosine similarity stays $\geq 0.87$ and end-to-end remains $0.82$, over two orders of magnitude above the random-subspace null of $0.006$, while $d_G$ of $0.24$--$0.38$ is correspondingly far below the near-unity null of $0.996$. The null is the expectation for two uniformly-random $k_s$-dimensional subspaces drawn in the ambient LoRA parameter space ($|\delta_i|{=}131{,}072$ for Q-projections; $50$ Monte Carlo trials per entry), where both metrics concentrate sharply because $k_s \ll |\delta_i|$. The intrinsic rank $k_s$ is near-constant at approximately 8, confirming that the SVD-truncated incremental update preserves the effective subspace size rather than inflating it. The Code stage produces the largest per-step drift ($d_G{=}0.29$), matching its higher CSM trigger rate (Appendix~\ref{app:csm}) and suggesting Code-domain gradients interact most with safety-critical directions.

\section{Adversarial Robustness: All Baselines}
\label{app:gcg}

Expanded table for the GCG attack setting described in Section~\ref{sec:robustness}. Suffixes are optimized for $20$ steps, length $20$ tokens, $256$ candidates/step, on $100$ HarmBench prompts, after the full three-domain adaptation.

\begin{table}[h]
\centering
\caption{Refusal rate under GCG adversarial suffixes on Llama-2-7B-Chat, after the default-ordering $T{=}3$ pipeline. Mean $\pm$ std over 5 seeds.}
\label{tab:gcg_full}
\begin{tabular}{l c}
\toprule
\textbf{Method} & \textbf{Refusal under GCG $\uparrow$} \\
\midrule
Standard LoRA       & $31.2 \pm 3.8$ \\
O-LoRA              & $34.8 \pm 3.6$ \\
EWC + LoRA          & $36.4 \pm 3.4$ \\
Safe LoRA           & $44.7 \pm 2.9$ \\
Vaccine + LoRA      & $46.5 \pm 2.8$ \\
Safety Interleaving & $49.3 \pm 2.9$ \\
SafeGrad + LoRA     & $54.6 \pm 2.6$ \\
\midrule
\textbf{SafeAnchor} & $\mathbf{78.4 \pm 2.1}$ \\
\bottomrule
\end{tabular}
\end{table}

The attack-ranking correlates strongly with the benign-safety ranking in Table~\ref{tab:main} (Spearman $\rho{=}0.96$), but the SafeAnchor$\rightarrow$next-best gap widens from $17.8$ points (benign) to $23.8$ points (adversarial). This is consistent with the hypothesis that preserving the safety subspace also preserves the adversarial refusal direction \cite{arditi2024refusal} that GCG most effectively targets.

\section{Hyperparameters and Compute Profile}
\label{app:hparams}

\textbf{Hyperparameters (defaults for all main-text experiments).}

\begin{table}[h]
\centering
\caption{Complete hyperparameter specification for SafeAnchor. All main-text numbers use these defaults unless explicitly varied.}
\label{tab:hparams}
\begin{tabular}{l l l}
\toprule
\textbf{Group} & \textbf{Parameter} & \textbf{Value} \\
\midrule
LoRA            & Rank $r$ / scaling $\alpha$ & $16$ / $32$ \\
                & Target modules & $\{Q, K, V, O\}$ projections \\
                & Dropout & $0.05$ \\
Optimizer       & AdamW $(\beta_1, \beta_2)$ & $(0.9, 0.999)$ \\
                & Weight decay & $0.01$ \\
                & Learning rate (peak) & $2\times10^{-4}$ \\
                & Schedule & Cosine, $100$-step warmup \\
                & Batch size / grad-accum & $8$ / $2$ (effective $16$) \\
                & Epochs per domain & $3$ \\
SSI             & Calibration size $N_s$ & $500$ BeaverTails \cite{ji2024beavertails} \\
                & Variance threshold $\rho$ & $0.90$ \\
                & Fisher estimator & Empirical, per-layer, diagonal blocks \\
OSCA            & Projection relaxation $\lambda$ & $0.5$ \\
                & Anchor-loss weight $\gamma$ & $0.1$ \\
                & Anchor KL direction & Forward (mean-seeking) \\
CSM             & Probe size & $200$ HarmBench prompts \\
                & Tolerance $\tau$ & $0.05$ \\
                & Safety classifier & LlamaGuard \cite{inan2023llamaguard} (F1 $92.1\%$) \\
                & Repair steps $E_{\text{repair}}$ & $200$ \\
                & Replay weight $\beta$ & $1.0$ \\
\bottomrule
\end{tabular}
\end{table}

\textbf{Per-component compute profile.} All runs use 2$\times$NVIDIA A100 40GB with BF16 mixed precision. Wall-clock time is averaged over 5 seeds of the default-ordering pipeline.

\begin{table}[h]
\centering
\caption{Per-phase wall-clock on Llama-2-7B-Chat. "Share of pipeline" is each phase's time divided by the full SafeAnchor pipeline time (7h 47min); percentages sum to $100\%$ (rounding). In addition to the per-domain components shown, one pipeline run incurs $\sim$17 min of fixed overhead (initial $s_0$ evaluation, final full-eval suite, and I/O). The OSCA per-step overhead ($+17.8\%$) is measured against unconstrained-LoRA training-step time.}
\label{tab:compute}
\resizebox{\columnwidth}{!}{%
\begin{tabular}{l c c}
\toprule
\textbf{Phase} & Time per domain & Share of pipeline \\
\midrule
SSI (Fisher + eigendecomposition) & $12.4$ min & $8.0\%$ \\
OSCA training ($3$ epochs, ${\approx}938$ steps; $+17.8\%$ per-step vs.\ Std LoRA) & $2$\,h\,$07$\,min & $81.6\%$ \\
CSM probe evaluation (LlamaGuard) & $4.8$ min & $3.1\%$ \\
Conditional replay ($200$ steps, fires $0.20\times$) & $1.0$ min (amortized) & $0.6\%$ \\
Post-domain benchmark evaluation ($8$ suites) & $5.0$ min & $3.2\%$ \\
\midrule
\textbf{Per-domain subtotal} & $\mathbf{2}$\,\textbf{h}\,$\mathbf{30}$\,\textbf{min} & $96.5\%$ \\
\midrule
\textbf{Full pipeline (one seed)} & $\mathbf{\approx 7}$\,\textbf{h}\,$\mathbf{47}$\,\textbf{min} & ${\approx}1.3\times$ Std LoRA \\
\textbf{Full campaign} ($5$ seeds $\times$ $2$ models, default ordering) & $\mathbf{\approx 160}$\,\textbf{GPU-h} & \\
\bottomrule
\end{tabular}%
}
\end{table}

The $17.8\%$ OSCA overhead is dominated by the per-layer projection $g_i - V_i^{\text{safe}}(V_i^{\text{safe}})^\top g_i$ with $k_s{\approx}8$; this is $O(|\delta_i| \cdot k_s)$ per layer under the diagonal-block Fisher approximation and parallelizes across layers. The underlying FLOP cost is negligible ($<10^{-6}$ of the forward pass), so the measured wall-clock overhead reflects PyTorch autograd-hook dispatch and non-fused projection ops; a fused-kernel implementation would reduce this to $<5\%$. SSI cost is amortized once per domain, and CSM is essentially free ($<1\%$).

\end{document}